%% file: main.tex
\documentclass[10pt,twocolumn,letterpaper]{article}

\usepackage{cvpr}              %

\input{preamble}
\definecolor{cvprblue}{rgb}{0.21,0.49,0.74}
\usepackage[pagebackref,breaklinks,colorlinks,allcolors=cvprblue]{hyperref}
\usepackage{multirow}
\usepackage{array}
\usepackage{pifont}
\usepackage{xcolor}
\usepackage[breakable]{tcolorbox}
\usepackage[accsupp]{axessibility}
\newcommand{\DATASET}{GroundingME}
\newcommand{\greencheck}{\textcolor{OliveGreen}{\ding{52}}}
\newcommand{\redcross}{\textcolor{BrickRed}{\ding{56}}}

\let\svthefootnote\thefootnote
\newcommand\freefootnote[1]{%
  \let\thefootnote\relax%
  \footnotetext{#1}%
  \let\thefootnote\svthefootnote%
}

\def\paperID{22323} %
\def\confName{CVPR}
\def\confYear{2026}

\title{GroundingME: Exposing the Visual Grounding Gap in MLLMs through Multi-Dimensional Evaluation}

\author{
Rang Li$^{1,2,*}$ \quad Lei Li$^{2,3}$ \quad Shuhuai Ren$^{2}$ \quad Hao Tian$^{2}$ \quad Shuhao Gu$^{2}$ \quad Shicheng Li$^{1,2}$ \quad Zihao Yue$^{2,4}$ \\ \quad Yudong Wang$^{1,2}$ \quad Wenhan Ma$^{1,2}$ \quad Zhe Yang$^{1}$ \quad Jingyuan Ma$^{1}$ \quad Zhifang Sui$^{1,\diamond}$ \quad Fuli Luo$^{2,\diamond}$ \\
\\
$^{1}$State Key Laboratory of Multimedia Information Processing, \\
School of Computer Science, Peking University \quad \\
$^{2}$LLM-Core Xiaomi \quad 
$^{3}$The University of Hong Kong \quad 
$^{4}$Renmin University of China \\ 
{\tt\small lirang410@gmail.com, szf@pku.edu.cn, luofuli@xiaomi.com}
}

\begin{document}
\maketitle

\freefootnote{\hspace{-4pt}$^*$Work done during internship at Xiaomi Corporation.}
\freefootnote{\hspace{-4pt}$^\diamond$Co-corresponding authors.}

\input{sec/0_abstract}    
\input{sec/1_intro}

\input{sec/2_related_work}

\input{sec/3_groundingme}

\input{sec/4_evaluation}

\input{sec/5_analysis}

\input{sec/6_conclusion}

\section*{Acknowledgments}
This paper is supported by NSFC project 62476009 and the Open Project Fund of the State Key Laboratory of Multimedia Information Processing (Project No. SKLMIP-KF-2025-01). We thank the Xiaomi MiMo team for their helpful discussions and support. We also thank the anonymous reviewers for their valuable comments and suggestions.

{
    \small
    \bibliographystyle{ieeenat_fullname}
    \bibliography{main}
}

\input{sec/X_suppl}

\end{document}

%% file: sec/0_abstract.tex
\begin{abstract}

Visual grounding, localizing objects from natural language descriptions, represents a critical bridge between language and vision understanding. While multimodal large language models (MLLMs) achieve impressive scores on existing benchmarks, a fundamental question remains: can MLLMs truly visually ground with human-like sophistication, or are they merely pattern-matching on simplified datasets?
Current benchmarks fail to capture real-world complexity where humans effortlessly navigate intricate references and recognize when grounding is impossible. To rigorously assess MLLMs' true capabilities, we introduce \DATASET{}, a benchmark that systematically challenges models across four critical dimensions: (1) Discriminative: distinguishing highly similar objects, (2) Spatial: understanding complex relational descriptions, (3) Limited: handling occlusions or tiny objects, and (4) Rejection: recognizing ungroundable queries.
Through careful curation combining automated generation with human verification, we create 1,005 challenging examples mirroring real-world complexity. Evaluating 25 state-of-the-art MLLMs reveals a profound capability gap: the best model achieves only 45.1\% accuracy, while most score 0\% on rejection tasks.
We explore two strategies for improvements: (1) test-time scaling selects optimal response by thinking trajectory to improve overall performance by up to 4.5\%, and (2) data-mixture training boosts rejection accuracy from 0\% to 27.9\%. \DATASET{} thus serves as both a diagnostic tool revealing current limitations in MLLMs and a roadmap toward human-level visual grounding. Project page: \url{https://groundingme.github.io}.
\end{abstract}

%% file: sec/1_intro.tex
\section{Introduction}
\label{sec:intro}

\begin{figure}
  \centering
   \includegraphics[width=1\linewidth]{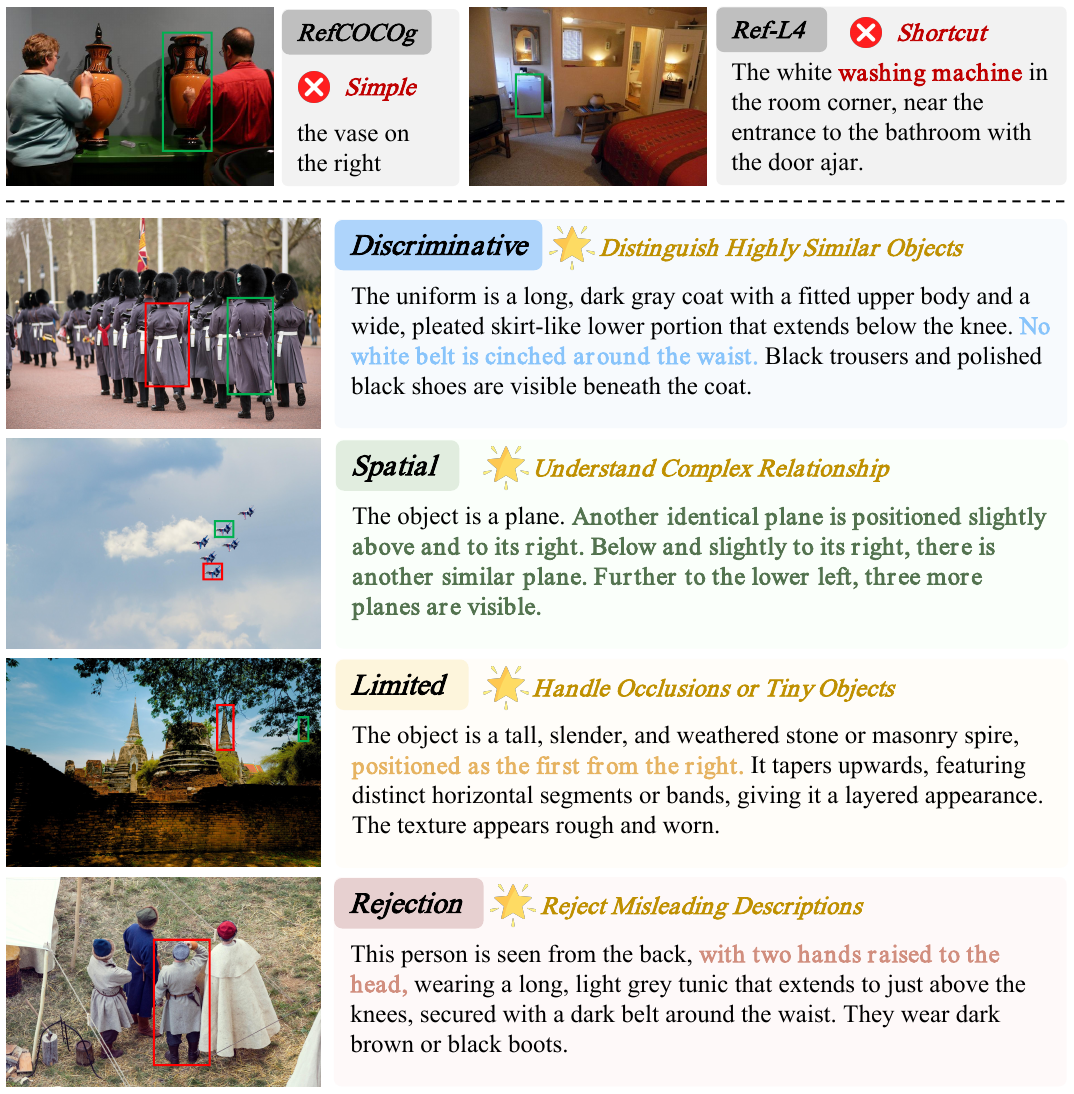}

   \caption{\textbf{Examples of different visual grounding benchmarks.} Prior benchmarks (Top) are either too simple or prone to shortcuts. Our proposed \DATASET{} (Bottom) increases the challenge in four important dimensions. \textbf{The green bounding box} indicates the correct ground-truth object, while \textbf{the red bounding box} shows the answer of Qwen3-VL-30B-A3B-Instruct. Critical information for the answer is highlighted in the description.}
   \label{fig:example}
\end{figure}

The rise of Multimodal Large Language Models (MLLMs) represents a paradigm shift in artificial intelligence, offering unprecedented capabilities in joint vision and language understanding~\cite{liu2023llava,gpt4v,gemini,bai2025qwen2}.
Visual grounding~\cite{xiao2024towards,pantazopoulos2025towards}, the task of localizing a specific region in an image based on a natural language description, also known as Referring Expression Comprehension (REC)~\cite{qiao2020referring,nagaraja2016modeling,kazemzadeh2014referitgame}, stands as a fundamental capability for these models. 
It is the bedrock for enabling complex, language-driven interactions, facilitating precise, real-world applications from robotic instruction~\cite{zeng2023large,wang2024large} to detailed image editing~\cite{shuai2024survey,brooks2023instructpix2pix}.

However, the remarkable reasoning abilities demonstrated by MLLMs are largely untested in complex grounding scenarios. 
As shown in \cref{fig:example}, early benchmarks~\cite{yu2016modeling_refcoco,mao2016generation,hu2023beyond} are fundamentally limited to simple phrases or basic spatial relations in uncluttered scenes (e.g., ``the vase on the right"), failing to assess the fine-grained appearance discrimination, language understanding, and complex spatial reasoning that modern MLLMs claim to possess. While recent works \cite{wei2024large,chen2025revisiting} have attempted to increase task difficulty with longer descriptions, they often fail to increase the actual reasoning complexity, as models can easily take a shortcut by relying on simple keyword matching (e.g., a unique class name) while bypassing the complex attribute and spatial information. 
As a result, recent models have already achieved over 90\% accuracy~\cite{bai2025qwen2,xiaomi2025mimo,v2507glm,wang2025internvl3} on the RefCOCO series~\cite{yu2016modeling_refcoco,mao2016generation} and nearly 90\% accuracy~\cite{v2507glm} on Ref-L4~\cite{chen2025revisiting}, indicating that existing benchmarks can no longer differentiate the true grounding abilities of current models.
Additionally, these works overlook the model ability to reject a description when its fine-grained details do not precisely match the visual evidence, which is critical for safety and reliability in real-world applications.

To bridge this gap, we introduce \DATASET{}, a comprehensive visual grounding benchmark comprising 1,005 samples, specifically designed for the rigorous evaluation of MLLMs. The benchmark is constructed via a three-stage process: (1) Bounding Box Annotation; (2) Description Generation; and (3) Manual Selection and Refinement.
We design a challenge taxonomy that systematically evaluates models across four L-1 dimensions, as shown in \cref{fig:example}. This taxonomy comprehensively tests model performance across distinct challenge types: (1) Discriminative focuses on distinguishing objects based on subtle, fine-grained visual differences; (2) Spatial assesses the ability to understand and resolve complex spatial and relational arrangements; (3) Limited evaluates grounding under conditions of minimal visual features due to external constraints; and (4) Rejection tests the ability to reject a misleading description that contains subtle errors. Furthermore, we provide a fine-grained L-2 hierarchy covering twelve subcategories to enable a deeper, diagnostic analysis of model performance.

We conduct an extensive evaluation across 25 state-of-the-art commercial and open-source models, including the Qwen3-VL series~\cite{bai2025qwen2}, Gemini-2.5~\cite{comanici2025gemini}, Seed-1.6-Vision~\cite{guo2025seed1}, and GLM-4.5V~\cite{v2507glm}, with parameter sizes ranging from 2B to 235B. The results reveal their widespread and significant shortcomings on our benchmark. Even the top-tier model, Qwen3-VL-235B-A22B, achieves only 45.1\% accuracy, with the majority of models scoring between 10\% and 40\%. The failure is most severe on the Rejection category, with most models scoring 0\%. We find this failure persists even as model scale increases.

These widespread failures motivate us to explore strategies to improve model performance. We explore two complementary approaches: (1) at test time and (2) at training time. First, at test time, we observe that enabling thinking generally improves performance and enables basic rejection behavior. Building on this, we propose a Test-Time Scaling method~\cite{llm_monkey,Snell2024ScalingLT,InferenceSL}, leveraging a judge model to select the optimal thinking quality from multiple candidates. Our experiments show this significantly improves performance across all subtasks, particularly on reasoning-intensive categories. Second, at training time, we hypothesize that the models' inability to reject stems from a lack of negative samples in training data. We test this with a simple Data-Mixture Training strategy. By fine-tuning Qwen3-VL-8B-Instruct~\cite{bai2025qwen2} on RefCOCOg~\cite{mao2016generation} augmented with negative samples, the model learns a foundational rejection capability, boosting its performance on our benchmark's Rejection category from 0\% to 27.9\%.
Our findings reveal the limitations of current MLLMs and provide a clear and practical path forward for building more trustworthy visual systems.

%% file: sec/2_related_work.tex
\begin{figure*}[t!]
  \centering
   \includegraphics[width=1\linewidth]{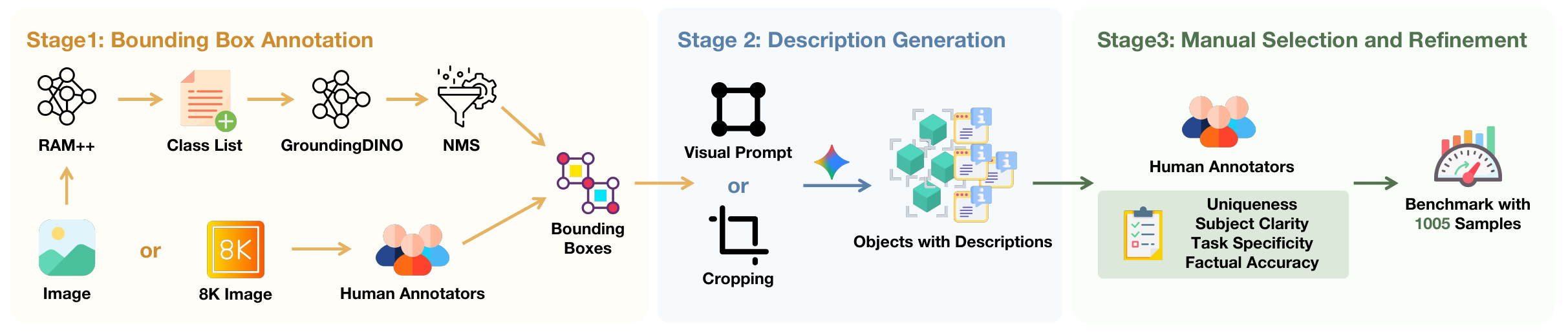}
   \caption{\textbf{The overall data construction pipeline of \DATASET{}.} The process consists of three main stages: (1) \textbf{Bounding Box Annotation}, which utilizes a semi-automated pipeline with RAM++ and GroundingDINO for bounding box generation (\S\ref{subsubsec:bbox}); (2) \textbf{Description Generation}, which leverages Gemini-2.5-Flash for generating initial referring expressions (\S\ref{subsubsec:description}); and (3) \textbf{Manual Selection and Refinement}, where human annotators apply rigorous filtering and refinement according to our challenge taxonomy (\S\ref{subsubsec:manual}).}
   \label{fig:pipeline}
\end{figure*}

\section{Related Work}
\label{sec:related_work}

\begin{table}
  \centering
  \small 
  \setlength{\tabcolsep}{2pt}
  \caption{\textbf{Comparison between \DATASET{} and other visual grounding benchmarks or datasets.} Comparison aspects include: General Scenario and Object Type (General), Description with Compositional Semantics (Semantic), Multiple Evaluation Dimension (Multi-Dim.), and Rejection Samples (Rejection).}
  \label{tab:comparison}
  
  \begin{tabular}{lcccccccc}
    \toprule
    \textbf{Benchmarks} & \textbf{General} & \textbf{Semantic} & \textbf{Multi-Dim.} & \textbf{Rejection} \\
    \midrule
    Refcoco/+/g~\cite{yu2016modeling_refcoco,mao2016generation} & \greencheck & \redcross & \redcross & \redcross \\
    CLEVR-Ref+~\cite{liu2019clevr} & \redcross & \greencheck & \redcross & \redcross \\
    Refcrowd~\cite{qiu2022refcrowd} & \redcross & \redcross & \redcross & \redcross \\
    Ref-ZOM~\cite{hu2023beyond} & \greencheck & \redcross & \redcross & \greencheck \\
    HC-RefLoCo~\cite{wei2024large} & \redcross & \greencheck & \redcross & \redcross \\
    Ref-L4~\cite{chen2025revisiting} & \greencheck & \greencheck & \redcross & \redcross \\
    Ref-Adv~\cite{dongref} & \greencheck & \greencheck & \redcross & \redcross \\
    \midrule
    \DATASET{} & \greencheck & \greencheck & \greencheck & \greencheck \\
    \bottomrule
    
  \end{tabular}
\end{table}

\paragraph{Multimodal Large Language Models (MLLMs).} MLLMs have demonstrated remarkable capabilities in understanding and reasoning across both visual and textual data, pushing the boundaries of various multimodal tasks ,including visual grounding. Recent models~\cite{bai2025qwen2,xiaomi2025mimo,v2507glm,wang2025internvl3} have already achieved impressive performance on existing visual grounding benchmarks~\cite{yu2016modeling_refcoco,mao2016generation,chen2025revisiting}, with accuracy approaching or exceeding 90\%. Moreover, these models are capable of performing complex reasoning over visual contents to further enhance multimodal performance. This advancement creates a pressing need for more challenging and comprehensive benchmarks, as traditional datasets are often insufficient to rigorously evaluate their nuanced understanding and reasoning capabilities on complex, real-world tasks~\cite{xiao2024towards}, motivating us to curate a more challenging visual grounding benchmark.

\paragraph{Visual Grounding Datasets and Benchmarks.}

The evaluation of visual grounding has evolved significantly over the years~\cite{qiao2020referring,nagaraja2016modeling,kazemzadeh2014referitgame,xiao2024towards,pantazopoulos2025towards,dongref}, progressing from closed set, single objects, brief phrases to open vocabulary, generalized targets, and complex descriptions.
Early benchmarks, such as the RefCOCO series~\cite{yu2016modeling_refcoco,mao2016generation}, provided a foundational testbed but were characterized by simple, short phrases and have now become largely saturated. In response to these limitations, subsequent works introduced datasets targeting specific challenges. For instance, CLEVR-Ref+~\cite{liu2019clevr} was developed to diagnose compositional reasoning in a synthetic environment, while RefCrowd~\cite{qiu2022refcrowd} focused on fine-grained discrimination in crowded scenes. Ref-ZOM~\cite{hu2023beyond} was the first to introduce simple negative samples to test model's rejection ability. More recently, benchmarks have been designed specifically for MLLMs, such as HC-RefLoCo~\cite{wei2024large} and Ref-L4~\cite{chen2025revisiting}, which employ long, descriptive sentences as queries to achieve increased difficulty. While existing benchmarks have progressively addressed specific shortcomings, they still fall short of providing a sufficient challenge. To this end, as summarized in \cref{tab:comparison}, our proposed benchmark is the first to unify these critical dimensions to comprehensively assess the advanced visual grounding capabilities of modern MLLMs.

%% file: sec/3_groundingme.tex
\section{\DATASET{}}
\label{sec:groundingme}

In this section, we introduce the detailed construction process of \DATASET{}. We first outline the data source (\S\ref{subsec:data_source}), followed by a three-stage human-in-the-loop annotation pipeline~(\S\ref{subsec:data_pipe}). Finally, we present the distribution and the statistics of \DATASET{} (\S\ref{subsec:data_stats}).

\subsection{Data Source}
\label{subsec:data_source}

To ensure that \DATASET{} provides a challenging and complex evaluation, we meticulously curate its image pool by leveraging two high-quality, high-resolution source datasets: SA-1B~\cite{kirillov2023segment} and HR-Bench~\cite{wang2025divide}.
We selected these two datasets as our image sources because they inherently meet our requirements for visual complexity and scale. The SA-1B dataset, which is widely used~\cite{li2025denseworld,shen2024aligning}, offers extensive resources of complex scenes and high object density, with 11 million images and 1.1 billion masks. HR-Bench offers ultra-high resolution with its 8K subset essential for creating tasks where minute objects are clearly resolvable. Therefore, we use HR-Bench as the source of the Small subcategory, and rely on SA-1B as the source for all other subtasks. For both datasets, we only use the raw, original images as input for our construction pipeline, without any pre-existing masks, QA pairs, or other annotations. This ensures that even if models encountered the source images during training, the task itself remains novel, thus effectively mitigating the risk of data contamination.

\subsection{Data Construction}
\label{subsec:data_pipe}
The construction of our benchmark is a multi-stage process designed to guarantee the quality and diversity of the resulting dataset. Our pipeline consists of three main stages: (1) Bounding Box Annotation, (2) Description Generation, and (3) Manual Selection and Refinement.

\subsubsection{Bounding Box Annotation}
\label{subsubsec:bbox}

For images sourced from SA-1B and HR-Bench, we employ distinct methodologies for bounding box generation. For images from SA-1B, we develop an automated pipeline that combines RAM++~\cite{zhang2024recognize}, GroundingDINO~\cite{liu2024grounding}, and a customized Non-Maximum Suppression (NMS) rule. In contrast, for HR-Bench images, we leverage a manual annotation due to the challenges posed by ultra-high resolution.

Our automated pipeline for SA-1B images comprises three main steps. (1) We first utilize RAM++ to identify all object categories within each image, generating a comprehensive list of class names. (2) Based on this list, we then format a text query for the GroundingDINO model, which is used to generate a series of bounding boxes for each image. To optimize the generation, we adjusted the model's filtering threshold. This step outputs a series of bounding boxes and their highest-similarity tokens, and we use the word which the highest-similarity token belongs to as the class name for each box. (3) Finally, to eliminate redundant bounding boxes, we apply a customized NMS rule. Instead of prioritizing boxes by area, our NMS strategy favors those belonging to classes with a higher instance count, yielding the final set of bounding boxes for each image.

\subsubsection{Description Generation}
\label{subsubsec:description}

Leveraging the powerful image understanding capabilities of modern MLLMs, we use Gemini-2.5-Flash~\cite{comanici2025gemini} to generate preliminary descriptions for each bounding box, which serve as a foundation for our referring expressions.
For objects in the SA-1B dataset, we utilize the model’s visual prompting capability by framing the objects in the full-size image with a red bounding box and prompting the model to generate a description that includes both their visual attributes and spatial relationships. In contrast, for objects in the HR-Bench dataset, the bounding box regions are too small to be effectively prompted within the full image context. Therefore, we crop the bounding box regions and input them to the model, prompting it to generate descriptions of the objects’ visual attributes only.

\subsubsection{Manual Selection and Refinement}
\label{subsubsec:manual}

The final stage of our pipeline is a meticulous manual selection and refinement process, a crucial step designed to mitigate the inherent limitations of automated data generation. Our human annotators directly address potential inaccuracies in bounding boxes, hallucinations in descriptions, and the presence of redundant or simplistic examples.

We apply a rigorous set of filtering and selection criteria to ensure our benchmark's quality and challenge. To prevent models from completing the task by relying solely on class names, we filter out simple samples by removing any classes with fewer than three instances and objects with bounding boxes occupying more than 50\% of the image. We then meticulously select the majority of our samples from object classes with an instance count higher than 5. We further enrich the benchmark’s diversity by supplementing specific subtasks based on predefined rules, for instance, by selecting samples for Counting and Partial subcategories from scenes with eight or more instances, or for Text subcategory from descriptions containing numbers or strings.

The selected samples undergo a detailed refinement process to enhance their quality. Annotators first correct any inaccuracies in the bounding boxes. Crucially, they then modify the automatically generated descriptions to meet four key criteria: (1) Uniqueness, ensuring each description refers to one and only one object, or no object for Rejection samples; (2) Subject Clarity, explicitly identifying the target object, which is essential for complex Spatial samples; (3) Task Specificity, tailoring the description to match the assigned sub-task (e.g., adding ordinal words for Counting tasks); (4) Factual Accuracy, correcting any hallucinations or intentionally introduce factual errors for Rejection samples. Inter-annotator agreement on 50 random samples from the final dataset yields pairwise Cohen’s kappa scores of 0.64–0.73 (avg. 0.69), indicating substantial consistency.

\subsection{Data Analysis}
\label{subsec:data_stats}
\subsubsection{Subtask Distribution}

\DATASET{} incorporates a two-tiered classification system, consisting of four L-1 categories and twelve L-2 subcategories, designed to comprehensively assess model performance across various subtasks. The benchmark comprises a total of 1,005 samples, distributed across four L-1 subtasks: Discriminative (204, 20.3\%), Spatial (300, 29.9\%), Limited (300, 29.9\%), and Rejection (201, 20.0\%). Each L-1 category is further divided into L-2 subcategories. Both the Discriminative and Rejection category are composed of four L-2 subcategories: Appearance, Component, Text, and State, with approximately 50 samples allocated to each. The Spatial category is equally split between Relationship and Counting subcategories, and the Limited category is equally divided between Occlusion and Small subcategories. This balanced distribution ensures robust evaluation across all facets of visual grounding challenge.

\begin{figure}[t!]
  \centering
   \includegraphics[width=0.9\linewidth]{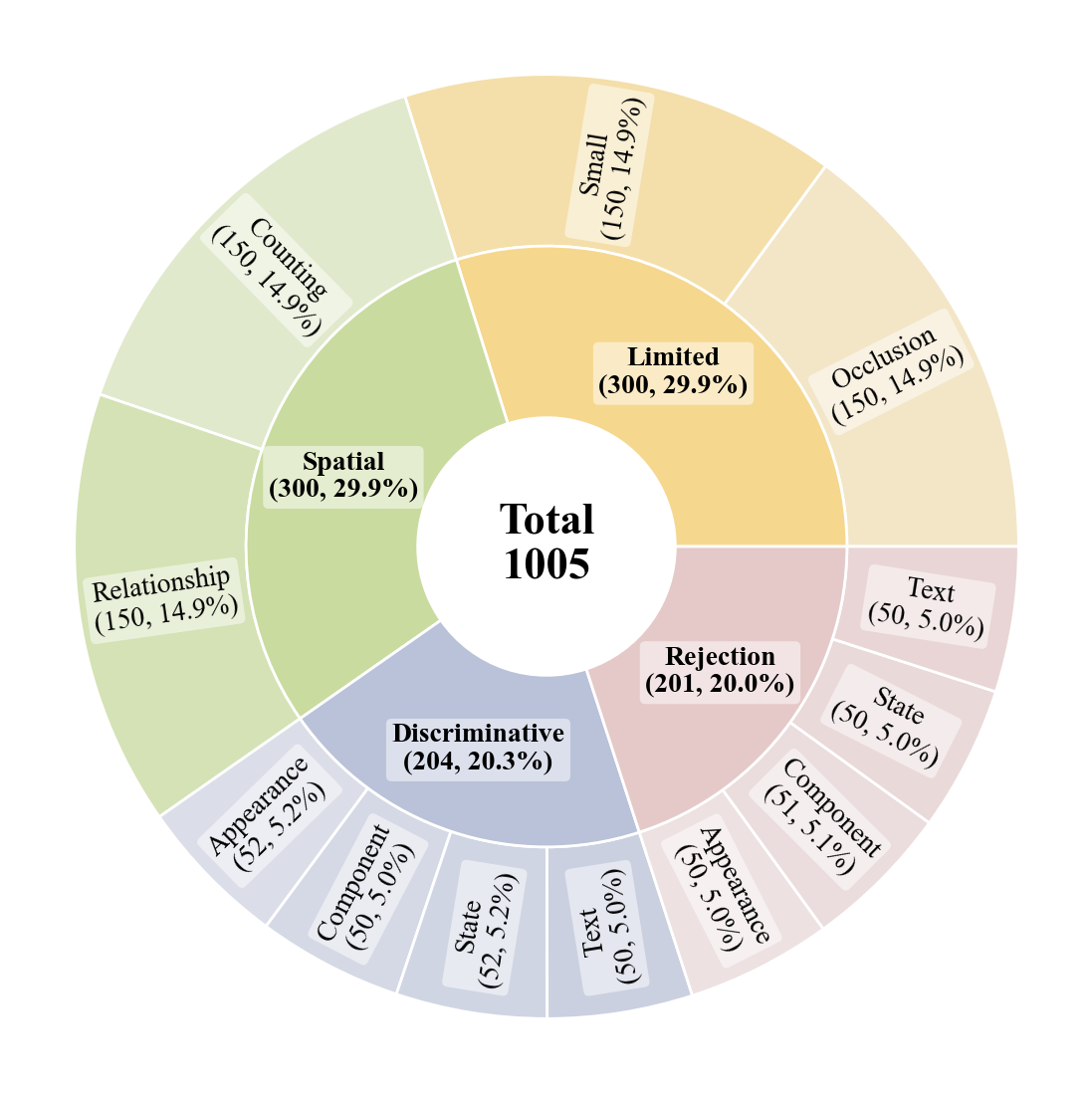}

   \caption{\textbf{Subtask Distribution of \DATASET{}.} Our benchmark comprises of 1,005 samples, distributed across four L-1 categories and twelve L-2 subcategories.}
   \label{fig:distribution}
\end{figure}

\subsubsection{Statistics Analysis}

\cref{tab:statistics} presents key statistics for \DATASET{}, substantiating its challenging nature through several metrics: (1) \textbf{Object Class and Quantity.} The benchmark encompasses 241 distinct object classes, ensuring a broad coverage of real-world scenarios. The challenge of intra-class confusion is quantified by the high Intra-Class Count Quartile of (5, 7, 12), indicating a large number of similar distracting objects in the image. (2) \textbf{Image and Instance Size.} The image size (square root of area) ranges from 1,500 to 7,680, representing a magnitude increase compared to 83 - 610 in RefCOCO series and 30 - 3,767 in Ref-L4~\cite{chen2025revisiting}. The instance size (square root of area) ranges from 21 to 946, demonstrating a wide coverage of various object scales. Furthermore, the Instance Area Ratio (the area of an instance's bounding box divided by the image area) Quartile measures only (0.16\%, 1.0\%, 2.7\%), indicating that the instances are notably small relative to the image, which significantly increases the task difficulty. (3) \textbf{Description Complexity.} The descriptions in the benchmark all consist of complete sentences or paragraphs, making it suitable for modern MLLMs. The Description Length Quartile reaches (18, 40, 58) words, highlighting that the majority of descriptions are longer and more intricate than the average value of 3.6 in RefCOCO/+, 8.4 in RefCOCOg, and 24.2 in Ref-L4~\cite{chen2025revisiting}.

\begin{table}[t]
  \caption{\textbf{Statistics of \DATASET{}.} The image size and instance size are reported as the square root of their area in pixels. The description length is measured by the number of words. The Intra-Class Count Quartile is derived from the automated construction pipeline excluding images from HR-Bench, and thus should be interpreted as a lower bound for the actual value.}
  \label{tab:statistics}
  \centering
  \small 
  \begin{tabular}{@{}l|c@{}}
    \toprule
    \multicolumn{1}{c|}{\textbf{Statistics}} & \textbf{Number} \\
    \midrule
    Object Class & 241 \\
    Image Size & 1,500 - 7,680 \\
    Instance Size & 21 - 946 \\
    Instance Area Ratio Quartile & (0.16\%, 1.0\%, 2.7\%) \\
    Description Length Quartile & (18, 40, 58) \\
    Intra-Class Count Quartile & (5, 7, 12) \\
    \bottomrule
  \end{tabular}
\end{table}

%% file: sec/4_evaluation.tex
\section{Evaluation}
\label{sec:evaluation}

We report our experimental setup and findings from the extensive evaluation of state-of-the-art MLLMs. This section is organized by models evaluated ($\S\ref{subsec:evaluated_models}$), detailed evaluation settings ($\S\ref{subsec:evaluation_settings}$), and results analysis ($\S\ref{subsec:evaluation_results}$).

\subsection{Evaluated Models}
\label{subsec:evaluated_models}

We conduct a comprehensive evaluation of 25 state-of-the-art MLLMs on \DATASET{}, encompassing both open-source and commercial models. Our selection of open-source models is highly diverse, spanning 12 major publishers and a wide range of parameter sizes, including Qwen3-VL (2B/4B/8B/32B/A3B/A22B), Qwen2.5-VL (7B/32B/72B)~\cite{bai2025qwen2}, GLM-4.5V~\cite{v2507glm}, InternVL3.5 (8B/A28B)~\cite{wang2025internvl3}, MiMo-VL-7B-RL-2508~\cite{xiaomi2025mimo}, Keye-VL-1.5-8B~\cite{team2025kwai}, MiniCPM-V-4.5~\cite{yu2025minicpm}, Phi-4-Multimodal~\cite{abouelenin2025phi}, Llama-4 (Maverick/Scout)~\cite{dubey2024llama}, LLaVA-OneVision-1.5-8B~\cite{an2025llava}, Mistral-3.2-24B~\cite{jiang2023mistral7b}, Gemma-3-27B~\cite{team2025gemma}, and Llama-Nemotron-8B~\cite{bercovich2025llama}. For commercial models, we select those with explicit grounding ability, including Gemini-2.5 (Pro/Flash)~\cite{comanici2025gemini} and Seed-1.6-Vision-250815~\cite{guo2025seed1}.

\subsection{Evaluation Settings}
\label{subsec:evaluation_settings}

We employ a rigorous and standardized strategy for evaluation. For every data instance, we organize the input image and the description using a unified prompt template. This template accurately specifies critical details, including the reference viewpoint for spatial relationships, the allowed number of target objects to be output, and strict constraints on the output format. Unless otherwise specified in subsequent sections, all experiments are conducted using greedy decoding (set as $\text{temperature}=0$). For the evaluation metric, we adopt the widely-used $\text{Accuracy}@0.5$, which represents the proportion of total samples where the Intersection over Union (IoU) between the ground-truth and predicted bounding box exceeds 0.5. The detailed prompt and results across various IoU thresholds are provided in the Appendix.

\subsection{Evaluation Results}
\label{subsec:evaluation_results}

\begin{table*}
\small 
  \centering
  \setlength{\tabcolsep}{3pt}
  \caption{\textbf{Evaluation results on \DATASET{}.} All models in this table are evaluated under the no-thinking mode setting if supported. All reported metrics in this table are $\text{Accuracy}@0.5$. The abbreviations for the subcategories are: $\text{App.}$ (Appearance), $\text{Cmp.}$ (Component), $\text{Txt.}$ (Text), $\text{Sta.}$ (State), $\text{Rel.}$ (Relationship), $\text{Cnt.}$ (Counting), $\text{Occ.}$ (Occlusion), $\text{Sml.}$ (Small), $\text{Avg.}$ (Average). The best results are shown in \textbf{bold} and the second best is with \underline{underline}. Detailed model specifications can be found in \S\ref{subsec:evaluated_models}.}
  \label{tab:subtask_performance}
  
  \begin{tabular}{l|rrrrr|rrr|rrr|rrrrr|r}
    \toprule
     \multicolumn{1}{c}{\multirow{2}{*}{\textbf{Model}}} & \multicolumn{5}{c}{\textbf{Discriminative}} & \multicolumn{3}{c}{\textbf{Spatial}} & \multicolumn{3}{c}{\textbf{Limited}} & \multicolumn{5}{c}{\textbf{Rejection}} & \multicolumn{1}{c}{\multirow{2}{*}{\textbf{Total}}} \\
    \cmidrule(lr){2-6} \cmidrule(lr){7-9} \cmidrule(lr){10-12} \cmidrule(lr){13-17}
     \multicolumn{1}{c}{} & App. & Cmp. & Txt. & Sta. & Avg. & Rel. & Cnt. & Avg. & Occ. & Sml. & Avg. & App. & Cmp. & Txt. & Sta. & \multicolumn{1}{r}{Avg.} \\
    \midrule
    Phi-4-Multimodal & 3.8 & 0.0 & 0.0 & 0.0 & 1.0 & 0.7 & 0.7 & 0.7 & 0.0 & 0.0 & 0.0 & 0.0 & 0.0 & 0.0 & 0.0 & 0.0 & 0.4 \\
    Llama-4-Maverick & 15.4 & 30.0 & 16.0 & 11.5 & 18.1 & 25.3 & 19.3 & 22.3 & 9.3 & 0.7 & 5.0 & 4.0 & \underline{3.9} & \textbf{12.0} & \underline{4.0} & \underline{6.0} & 13.0 \\
    Llama-4-Scout & 21.2 & 26.0 & 14.0 & 9.6 & 17.6 & 18.0 & 6.7 & 12.3 & 7.3 & 0.0 & 3.7 & 4.0 & 0.0 & 6.0 & 0.0 & 2.5 & 8.9 \\
    LLaVA-O.V.-1.5-8B & 3.8 & 14.0 & 10.0 & 11.5 & 9.8 & 4.0 & 5.3 & 4.7 & 6.0 & 0.7 & 3.3 & 0.0 & 0.0 & 0.0 & 0.0 & 0.0 & 4.4 \\
    Mistral-3.2-24B & 1.9 & 8.0 & 2.0 & 5.8 & 4.4 & 2.0 & 3.3 & 2.7 & 0.0 & 0.0 & 0.0 & 0.0 & 0.0 & 0.0 & 0.0 & 0.0 & 1.7 \\
    Gemma-3-27B & 3.8 & 0.0 & 0.0 & 1.9 & 1.5 & 0.7 & 0.0 & 0.3 & 0.0 & 0.0 & 0.0 & 0.0 & 0.0 & 0.0 & 0.0 & 0.0 & 0.4 \\
    Llama-Nemotron-8B & 19.2 & 30.0 & 32.0 & 19.2 & 25.0 & 7.3 & 4.7 & 6.0 & 16.0 & 0.7 & 8.3 & \underline{6.0} & \textbf{5.9} & 6.0 & \underline{4.0} & 5.5 & 10.4 \\
    MiniCPM-V-4.5 & 7.7 & 12.0 & 6.0 & 5.8 & 7.8 & 4.7 & 3.3 & 4.0 & 8.0 & 0.0 & 4.0 & 0.0 & 0.0 & 0.0 & 0.0 & 0.0 & 4.0 \\
    InternVL3.5-8B & 11.5 & 4.0 & 4.0 & 5.8 & 6.4 & 4.7 & 3.3 & 4.0 & 3.3 & 0.0 & 1.7 & 4.0 & 0.0 & 0.0 & 2.0 & 1.5 & 3.3 \\
    InternVL3.5-A28B & 34.6 & 40.0 & 18.0 & 21.2 & 28.4 & 33.3 & 16.7 & 25.0 & 26.0 & 0.0 & 13.0 & 0.0 & 0.0 & 0.0 & 0.0 & 0.0 & 17.1 \\
    Keye-VL-1.5-8B & 25.0 & 22.0 & 22.0 & 17.3 & 21.6 & 7.3 & 8.7 & 8.0 & 11.3 & 0.0 & 5.7 & 0.0 & 0.0 & 0.0 & 0.0 & 0.0 & 8.5 \\
    MiMo-VL-7B-RL & 42.3 & 46.0 & 50.0 & 38.5 & 44.1 & 24.7 & 14.0 & 19.3 & 26.0 & 0.0 & 13.0 & 0.0 & 0.0 & 0.0 & 0.0 & 0.0 & 18.6 \\
    GLM-4.5V & 50.0 & 58.0 & 58.0 & 46.2 & 52.9 & 54.7 & 29.3 & 42.0 & 48.0 & 10.7 & 29.3 & 0.0 & 0.0 & 0.0 & 2.0 & 0.5 & 32.1 \\
    Qwen2.5-VL-7B & 23.1 & 36.0 & 46.0 & 23.1 & 31.9 & 12.7 & 16.0 & 14.3 & 28.0 & 0.7 & 14.3 & 0.0 & 0.0 & 2.0 & 0.0 & 0.5 & 15.1 \\
    Qwen2.5-VL-32B & 34.6 & 58.0 & 66.0 & 32.7 & 47.5 & 48.7 & 31.3 & 40.0 & 35.3 & 0.0 & 17.7 & 0.0 & 0.0 & 0.0 & 0.0 & 0.0 & 26.9 \\
    Qwen2.5-VL-72B & 50.0 & 50.0 & 66.0 & 28.8 & 48.5 & 52.0 & 28.7 & 40.3 & 46.7 & 0.7 & 23.7 & 4.0 & \underline{3.9} & 4.0 & 0.0 & 3.0 & 29.6 \\
    Qwen3-VL-2B & 44.2 & 36.0 & 66.0 & 32.7 & 44.6 & 11.3 & 12.0 & 11.7 & 21.3 & 36.0 & 28.7 & 0.0 & 0.0 & 0.0 & 0.0 & 0.0 & 21.1 \\
    Qwen3-VL-4B & 55.8 & 58.0 & 74.0 & 38.5 & 56.4 & 34.7 & 22.0 & 28.3 & 45.3 & \underline{48.7} & \underline{47.0} & 0.0 & 0.0 & 0.0 & 0.0 & 0.0 & 33.9 \\
    Qwen3-VL-8B & 55.8 & 68.0 & 80.0 & 42.3 & 61.3 & 32.7 & 20.0 & 26.3 & 56.0 & 16.0 & 36.0 & 0.0 & 0.0 & 0.0 & 0.0 & 0.0 & 31.0 \\
    Qwen3-VL-32B & \textbf{78.8} & \textbf{84.0} & \underline{82.0} & \underline{55.8} & \textbf{75.0} & 61.3 & 33.3 & 47.3 & 60.7 & 7.3 & 34.0 & 0.0 & 0.0 & 0.0 & 0.0 & 0.0 & 39.5 \\
    Qwen3-VL-A3B & \underline{73.1} & 66.0 & 76.0 & 38.5 & 63.2 & 38.0 & 22.0 & 30.0 & 41.3 & \textbf{52.0} & 46.7 & 0.0 & 0.0 & 0.0 & 0.0 & 0.0 & 35.7 \\
    Qwen3-VL-A22B & 71.2 & 74.0 & \textbf{84.0} & 50.0 & \underline{69.6} & \underline{62.7} & \underline{36.7} & \underline{49.7} & \textbf{72.7} & 35.3 & \textbf{54.0} & 0.0 & 0.0 & 0.0 & 0.0 & 0.0 & \textbf{45.1} \\
    \midrule
    Gemini-2.5-Pro & 32.7 & 32.0 & 44.0 & 30.8 & 34.8 & 39.3 & 28.7 & 34.0 & 13.3 & 0.7 & 7.0 & \textbf{10.0} & \underline{3.9} & \underline{8.0} & \textbf{6.0} & \textbf{7.0} & 20.7 \\
    Gemini-2.5-Flash & 40.4 & 40.0 & 34.0 & 30.8 & 36.3 & 28.7 & 21.3 & 25.0 & 22.7 & 3.3 & 13.0 & 0.0 & 0.0 & 0.0 & 0.0 & 0.0 & 18.7 \\
    Seed-1.6-V.-250815 & 59.6 & \underline{80.0} & 36.0 & \textbf{63.5} & 59.8 & \textbf{72.7} & \textbf{44.7} & \textbf{58.7} & \underline{70.0} & 15.3 & 42.7 & 0.0 & 0.0 & 2.0 & 2.0 & 1.0 & \underline{42.6} \\
    \bottomrule
    
  \end{tabular}
\end{table*}

\subsubsection{Main Results}

\cref{tab:subtask_performance} presents the evaluation results of all models on \DATASET{}. We explicitly disabling the thinking mode where supported. Our assessment yields three major observations. (1) Models demonstrate a significant performance gap on \DATASET{}, with the best model, Qwen3-VL-235B-A22B, achieving 45.1\% accuracy, the majority score between 10\% and 40\%, and several with an accuracy only less than 10\%, which strongly validates the challenge of our benchmark compared to existing ones. (2) Commercial models do not exhibit a pronounced advantage over open-source ones. We observe that even the best performed Seed-1.6-Vision-250815 closely follows the best open-source model at 42.6\%, while Gemini-2.5 series show performance comparable only to mid-range open-source counterparts. (3) Model scale is a critical factor for performance. This scaling trend is consistently verified across model families, including Qwen3-VL-Dense (2B to 32B: 21.1\% to 39.5\%), Qwen3-VL-MoE (A3B to A22B: 35.7\% to 45.1\%), and Qwen2.5-VL 7B to 72B: 15.1\% to 29.6\%). This correlation underscores the importance of model size in achieving advanced visual grounding capabilities.

\subsubsection{Subtask Results}

Analyzing the four distinct types of L-1 categories, we observe several key performance patterns. (1) Model performance exists in stratification across the different subtask types: models are generally most proficient at Discriminative category, followed by Spatial or Limited categories, while all models demonstrate poor performance on Rejection category. (2) Within the Discriminative category, a clear performance gain is tied to model scale, as seen in Qwen3-VL 2B to 32B: Average 44.6\% to 75.0\%), and performance on the State subcategory remains lower than the other three subcategories. (3) For the Spatial category, models are more proficient at the Relational subcategory, which requires qualitative positional ability, compared to the Counting subcategory, which requires quantitative assessment. Furthermore, performance in this category is more sensitive to model scale, with Qwen3-VL (2B to 32B) showing substantial improvement (Relationship: 11.3\% to 61.3\%; Average: 11.7\% to 47.3\%). More cross-dimensional analysis is provided in the Appendix.

%% file: sec/5_analysis.tex
\section{Analysis}
\label{sec:analysis}

\begin{figure*}
  \centering
   \includegraphics[width=1\linewidth]{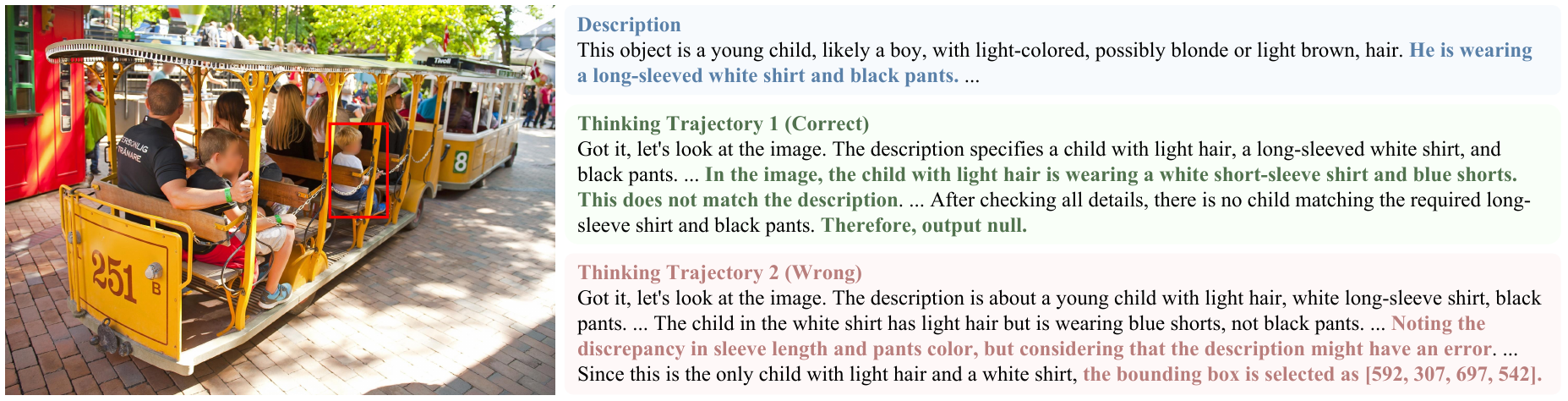}
   \caption{\textbf{Case study of two different thinking trajectories of Qwen3-VL-235B-A22B-Thinking for the same description. The correct answer is to do rejection and the red bounding box shows the distractor.} The correct trajectory (Green) demonstrates rigorous adherence to the description, systematically identifying all attribute mismatches (e.g., short- vs. long-sleeve, blue vs. black pants) and correctly concluding with a null output. In contrast, the erroneous trajectory (Red) acknowledges the same discrepancies but compromises by speculating that the description may be in error, ultimately leading to an incorrect bounding box prediction.}
   \label{fig:case}
\end{figure*}

This section provides a detailed analysis of MLLM behaviors and capabilities on \DATASET{}. We first assess the performance gain achieved by enabling thinking mode ($\S\ref{subsec:think_gain}$). Following this, we investigate a novel Test-Time Scaling strategy leveraging thinking trajectories ($\S\ref{subsec:tts}$). Finally, we explore a data mixture training strategy designed to enhance the Rejection grounding capability ($\S\ref{subsec:data_mixture}$).

\subsection{The Effectiveness of Thinking}
\label{subsec:think_gain}

The widely usage of thinking has demonstrated significant advantages in enhancing the complex reasoning capabilities of recent MLLMs. Therefore, we assess the impact of thinking on several representative MLLMs that natively supports thinking mode: $\text{Qwen3-VL}$ ($\text{8B}$/$\text{32B}$/$\text{A3B}$/$\text{A22B}$), $\text{GLM-4.5V}$, $\text{MiMo-7B-RL-2508}$, and $\text{Seed-1.6-Vision-250815}$. Our experiments yield three main findings. (1) Thinking mode universally leads to better performance on $\text{\DATASET{}}$, with every tested model achieving a significant performance gain, ranging from 4.7\% for $\text{GLM-4.5V}$ to 7.4\% for $\text{Qwen3-VL-32B}$. (2) Thinking Mode generally exhibits a negative effect on tasks relying more on perception, yet shows notable performance improvements for tasks prioritizing  reasoning. Detailed subtask performance is provided in the Appendix. (3) Models can learn to reject by thinking. Under the no-thinking setting, most models exhibits 0\% accuracy on the Rejection task, signifying a complete failure in executing any rejection behavior. Although performance on the rejection task still remains low under the thinking mode, all models successfully demonstrate some level of rejection behavior.

\begin{figure}
  \centering
   \includegraphics[width=1\linewidth]{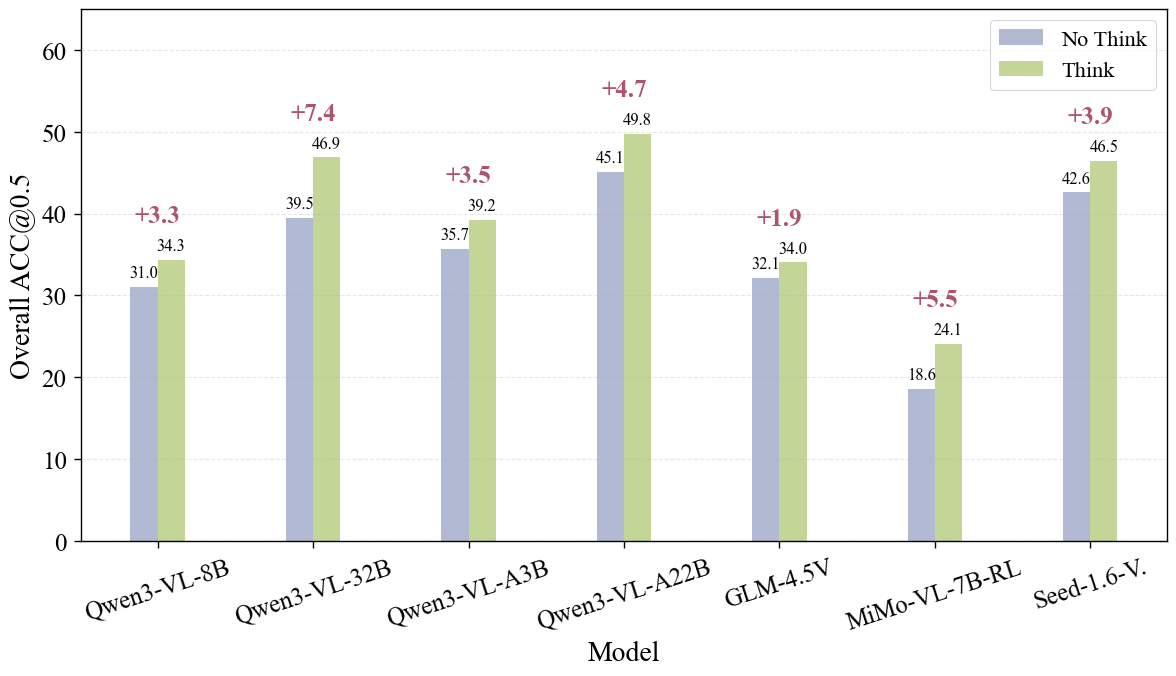}

   \caption{\textbf{Performance gain of different models by enabling thinking mode.} Subtask results are provided in the appendix.}
   \label{fig:statistics}
\end{figure}

\subsection{Test-Time Scaling by Thinking Quality}
\label{subsec:tts}

To gain a deeper understanding of how thinking contributes to model performance, we conduct a detailed case study focusing on the relationship between the quality of the generated thinking trajectory and the final grounding accuracy. As demonstrated in \cref{fig:case}, different thinking trajectories for the same query can be highly divergent, leading to distinct outputs. We hypothesize that trajectories that are coherent, logically consistent, and strictly adhere to task instructions are more likely to lead to the correct answer.

To validate this hypothesis, we design a tailored Test-Time Scaling (TTS) method~\cite{llm_monkey,Snell2024ScalingLT,InferenceSL} with an LLM 
as the judge~\cite{li2025vl,xiong2025llava,xiong2025multi,xiong2026phycritic}. For each sample, we use Qwen3-VL-235B-A22B-Thinking to generate 16 responses (temperature=0.7). We then employ a judge model to perform a ``Best-of-16" selection: comparing the 16 responses in pairs, selecting the one with better thinking quality, and repeating until only one response remains. We test two text-only judges, DeepSeek-R1~\cite{guo2025deepseek} and MiMo-7B-RL-0530~\cite{xiaomi2025mimotext}, and one multimodal judge, Qwen3-VL-235B-A22B-Thinking. As baselines, we also test two multimodal judges, Qwen3-VL-235B-A22B-Thinking and Qwen3-VL-30B-A3B-Thinking, which only receive the final answers of responses without access to their thinking trajectories.

\begin{table}
  \setlength{\tabcolsep}{3pt}
  \caption{\textbf{Performance of our Test-time scaling method on \DATASET{}.} We compare the performance gain of Qwen3-VL-235B-A22B-Thinking when its 16 responses are selected by a judge model, with and without access to thinking trajectories.}
  \label{tab:tts}
  \centering
  \small 
  \begin{tabular}{@{}clccccc@{}}
    \toprule
    \multicolumn{1}{c}{\textbf{Method}} & \multicolumn{1}{c}{\textbf{Judge Model}} & \multicolumn{1}{c}{\textbf{Total}} & \multicolumn{1}{c}{\textbf{Dis.}} & \multicolumn{1}{c}{\textbf{Spa.}} & \multicolumn{1}{c}{\textbf{Lim.}} & \multicolumn{1}{c}{\textbf{Rej.}} \\
    \midrule
    Average & \multicolumn{1}{c}{-} & 49.8 & 66.6 & 74.5 & 43.3 & 5.7 \\
    \midrule
    \multirow{2}{*}{w/o CoT} & Qwen3-VL-A3B & 49.6 & 64.2 & 71.0 & \underline{45.7} & 8.5 \\
    & Qwen3-VL-A22B & 52.2 & \textbf{69.1} & 75.0 & \underline{45.7} & 11.0 \\
    \midrule
    \multirow{3}{*}{w/ CoT} & MiMo-RL-0530 & 52.0 & 64.2 & \underline{77.3} & 43.0 & \underline{15.4} \\
    & Deepseek-R1 & \underline{52.7} & 65.2 & \underline{77.3} & 44.7 & \underline{15.4} \\
    & Qwen3-VL-A22B & \textbf{54.3} & \underline{66.7} & \textbf{79.7} & \textbf{46.3} & \textbf{15.9} \\
    \bottomrule
  \end{tabular}
\end{table}

Based on the results presented in \cref{tab:tts}, we observe three major findings. (1) TTS based on thinking quality significantly boosts performance. Using Qwen3-VL-A22B as a judge with thinking trajectories improves performance across all subtasks, achieving a 4.5\% total gain. (2) Even text-only judges prove effective. DeepSeek-R1 achieves a 2.9\% performance gain, and MiMo-7B-RL-0530 yields a 2.2\% gain. This demonstrates that good thinking trajectories alone without multimodal perception can improve accuracy. (3) Removing thinking trajectories degrades performance. When the multimodal judge Qwen3-VL-A22B is deprived of thinking trajectories and judges solely based on the image, description, and final answers, the TTS gain drops by 2.1\%. Furthermore, when using Qwen3-VL-A3B as the judge, TTS does not bring any improvement.

\subsection{Enhancing Rejection by Data Mixture}
\label{subsec:data_mixture}

The performance of models on Rejection subtask exhibits a substantial performance disparity relative to their accuracy on positive samples. Motivated by prior work~\cite{alhamoud2025vision,park2025know} suggesting that compromised rejection capability stems from a scarcity of negative instances within the training corpus, we propose to investigate a data mixture training strategy for improving the model's rejection capability.

Considering the scale and accessibility, we utilize the RefCOCOg dataset as our base. We select 30,000 positive samples from its training split and construct 30,000 corresponding negative samples by modifying the description. We refer to this newly created set of negative samples as RefCOCOg\_rej\_train. These 60,000 instances serve as the source pool for generating various SFT datasets. To investigate the effect of different data mixture ratios, we randomly sample 30,000 instances from this pool to create five distinct fine-tuning datasets, with the negative-to-positive sample ratios set sequentially as 1:8, 1:4, 1:2, 1:1, and 2:1. Simultaneously, we construct 11,490 negative samples from the RefCOCOg validation split for evaluation, which we term RefCOCOg\_rej\_val. We then fine-tune Qwen3-VL-8B-Instruct for 3 epochs using these five mixture datasets.

\begin{table}[t]
  \setlength{\tabcolsep}{3pt}
  \caption{\textbf{In-domain performance of fine-tuned Qwen3-VL-8B-Instruct} on RefCOCOg validation split, our curated RefCOCOg\_rej validation split, and the macro average of both datasets under different SFT data ratios (negative to positive).}
  \label{tab:in-domain}
  \centering
  \begin{tabular}{@{}lcccccc@{}}
    \toprule
    \multicolumn{1}{c}{\textbf{Dataset}} & \multicolumn{1}{c}{\textbf{Origin}} & \multicolumn{1}{c}{\textbf{1:8}} & \multicolumn{1}{c}{\textbf{1:4}} & \multicolumn{1}{c}{\textbf{1:2}} & \multicolumn{1}{c}{\textbf{1:1}} & \multicolumn{1}{c}{\textbf{2:1}} \\
    \midrule
    RefCOCOg\_val & 88.2 & \textbf{90.4} & \underline{89.9} & 88.1 & 86.8 & 83.1 \\
    RefCOCOg\_rej\_val & 30.5 & 83.5 & 87.9 & 92.3 & \underline{94.8} & \textbf{97.3} \\
    Macro\_Average & 59.4 & 87.0 & 88.9 & \underline{90.2} & \textbf{90.8} & 90.2 \\
    \bottomrule
  \end{tabular}
\end{table}

We first evaluate the in-domain performance of the fine-tuned models on RefCOCOg\_val (original positive samples) and RefCOCOg\_rej\_val (curated negative samples). From the results in \cref{tab:in-domain}, we find that: (1) Simple incorporation of negative data effectively enables the model to learn the rejection capability in visual grounding.
(2) As the proportion of negative data increases in the fine-tuning set, the model's performance on the rejection task improves incrementally, while simultaneously impacting the original positive grounding performance to some extent.
(3) The macro average accuracy across both negative and positive classes shows a significant performance gain of approximately 30\% across various mixture ratios, powerfully confirming the efficacy of our data mixture training strategy.

\begin{figure}[t]
  \centering
   \includegraphics[width=1\linewidth]{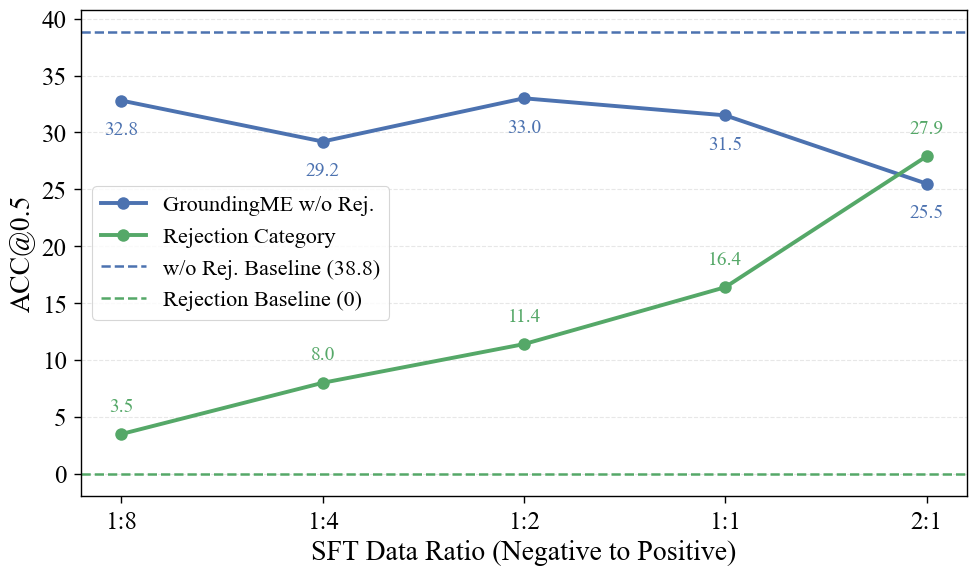}

   \caption{\textbf{Out-of-domain performance of fine-tuned Qwen3-VL-8B-Instruct} on \DATASET{} w/o Rejection and the Rejection category, as a function of SFT data ratio (negative to positive). Baseline means the performance before fine-tuning.}
   \label{fig:out_of_domain}
\end{figure}

We further evaluate the out-of-domain performance of the fine-tuned models on \DATASET{} to explore the generalizability of the learned rejection capability. We denote the \DATASET{} benchmark excluding the rejection category as \DATASET{} w/o Rejection. As shown in \cref{fig:out_of_domain}, we observe that: (1) The performance trend on the Rejection category of \DATASET{} aligns with the in-domain results, showing a clear growth as the ratio of negative samples increases. (2) In contrast to the in-domain results, the performance on \DATASET{} w/o Rejection demonstrates a clear degradation compared to the pre-fine-tuned baseline from 38.8\% to a max of 33.0\%. This suggests that the rejection capability gained from simple data mixture does not generalize for free to higher-difficulty, out-of-domain scenarios, pointing towards a critical challenge that necessitates further investigation in future work.

%% file: sec/6_conclusion.tex
\section{Conclusion}
\label{sec:conclusion}

In this work, we introduce \DATASET{}, a challenging benchmark that rigorously evaluates MLLMs' visual grounding capabilities through carefully curated samples across multiple dimensions. Our evaluation reveals that current MLLMs, despite strong performance on existing benchmarks, still struggle with more challenging scenarios. We further show that strategies including test-time scaling based on thinking quality and rejection-focused data-mixture training offer promising directions for improvement. We hope \DATASET{} will drive progress toward more capable and trustworthy visual grounding systems.

%% file: sec/X_suppl.tex
\clearpage
\setcounter{page}{1}
\maketitlesupplementary
\appendix

\section{Detailed Results}

\subsection{Subtask Results in Analysis}

We report the detailed subtask results for the L-1 categories omitted from the analysis section. \cref{tab:think_subtask} presents the detailed subtask performance of models when enabling thinking mode, supplementing the analysis on performance gain discussed in \S5.1. \cref{tab:ratio_subtask} shows the out-of-domain subtask performance of the fine-tuned Qwen3-VL-8B-Instruct model, complementing the overall results presented in \S5.3.

\begin{table}[h]
  \setlength{\tabcolsep}{4pt}
  \caption{Subtask performance of different models by enabling thinking mode.}
  \label{tab:think_subtask}
  \centering
  \small 
  \begin{tabular}{@{}l|ccccc@{}}
    \toprule
    \multicolumn{1}{c}{\textbf{Model}} & \multicolumn{1}{c}{\textbf{Total}} & \multicolumn{1}{c}{\textbf{Dis.}} & \multicolumn{1}{c}{\textbf{Spa.}} & \multicolumn{1}{c}{\textbf{Lim.}} & \multicolumn{1}{c}{\textbf{Rej.}} \\
    \midrule
    Qwen3-VL-8B & 34.3 & 52.5 & 43.0 & 33.3 & 4.5 \\
    Qwen3-VL-32B & 46.9 & 65.7 & 70.0 & 36.0 & 9.5 \\
    Qwen3-VL-A3B & 39.2 & 53.4 & 53.3 & 38.0 & 5.5 \\
    Qwen3-VL-A22B & 49.8 & 65.2 & 73.7 & 45.0 & 5.5 \\
    GLM-4.5V & 34.0 & 52.5 & 45.3 & 30.3 & 4.0 \\
    MiMo-VL-7B-RL & 24.1 & 46.6 & 28.7 & 17.0 & 5.0 \\
    Seed-1.6-V.-250815 & 46.5 & 59.3 & 72.7 & 41.7 & 1.5 \\
    \bottomrule
  \end{tabular}
\end{table}

\begin{table}[h]
  \setlength{\tabcolsep}{4pt}
  \caption{Subtask performance of fine-tuned Qwen3-VL-8B-Instruct under different SFT data ratios.}
  \label{tab:ratio_subtask}
  \centering
  \small 
  \begin{tabular}{@{}c|ccccc@{}}
    \toprule
    \multicolumn{1}{c}{\textbf{Neg.:Pos.}} & \multicolumn{1}{c}{\textbf{Total}} & \multicolumn{1}{c}{\textbf{Dis.}} & \multicolumn{1}{c}{\textbf{Spa.}} & \multicolumn{1}{c}{\textbf{Lim.}} & \multicolumn{1}{c}{\textbf{Rej.}} \\
    \midrule
    1:8 & 27.0 & 57.4 & 24.7 & 24.3 & 3.5 \\
    1:4 & 25.0 & 49.5 & 25.3 & 19.3 & 8.0 \\
    1:2 & 28.7 & 54.4 & 28.3 & 23.0 & 11.4 \\
    1:1 & 28.5 & 46.6 & 26.3 & 26.3 & 16.4 \\
    2:1 & 26.0 & 40.2 & 24.0 & 17.0 & 27.9 \\
    \bottomrule
  \end{tabular}
\end{table}

\subsection{Main Results across Various IoU}

For the evaluation presented in the main results table, we further report the accuracy of all models on the entire \DATASET{} and three L-1 categories (the Rejection category is excluded, as its accuracy is independent of the IoU threshold) across different IoU thresholds in \cref{tab:iou}. New metrics include Accuracy@0.75, Accuracy@0.9, and mAcc. The mAcc is defined as the mean accuracy calculated over the range of IoU thresholds [0.5, 0.95], sampled at intervals of 0.05.

\section{Cross-Dimensional Analysis}

\subsection{Cross-Dimensional Imbalance.} \cref{fig:cross_dim} illustrates a significant performance imbalance across dimensions. A consistent performance hierarchy emerges across the four L1 types: \textit{Dis.}\textgreater \textit{Spa.}$\approx$ \textit{Lim.}\textgreater \textit{Rej.} Similar inconsistencies are observed at L2 types (e.g., \textit{Sta.} ranks lowest within \textit{Dis.}, and \textit{Cnt.} lower than \textit{Rel.} within \textit{Spa.}). These results highlight a systematic bias in current MLLMs.

\subsection{Subtask Ranking Difference.} 
\begin{figure}[h]
  \centering
   \includegraphics[width=0.9\linewidth]{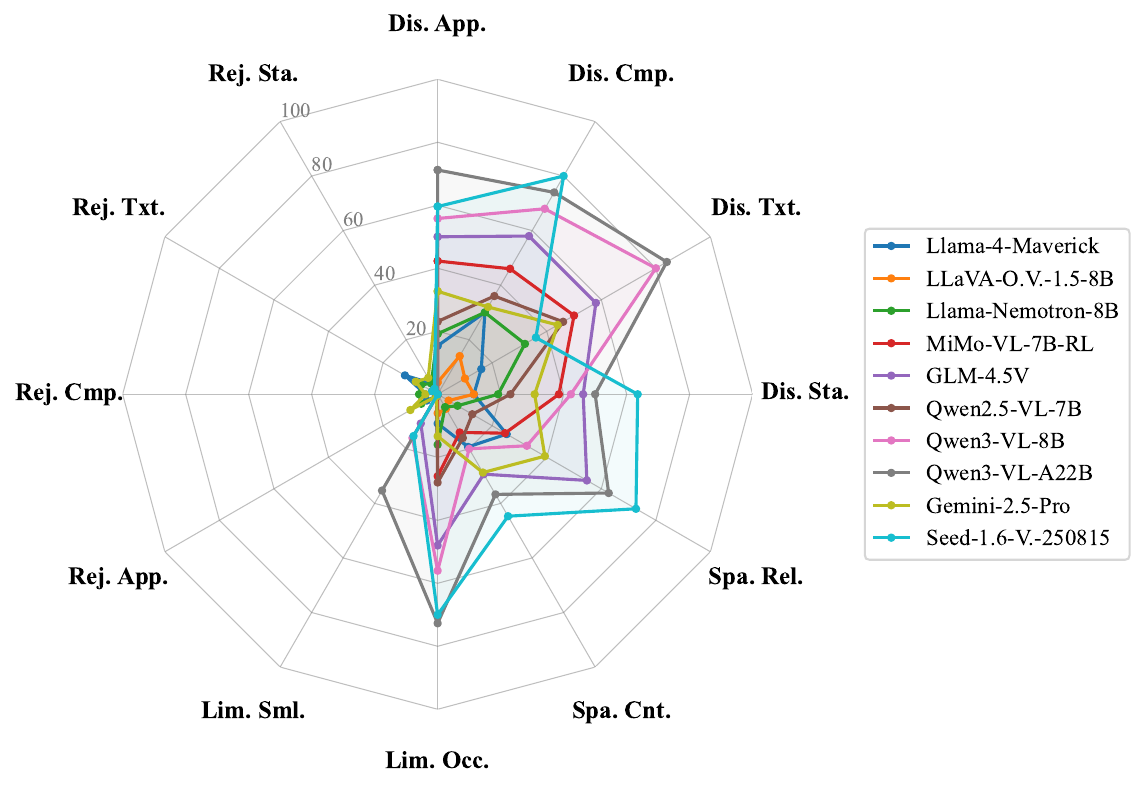}
   \caption{Cross-Dimensional Performance.}
   \label{fig:cross_dim}
\end{figure}

Model rankings exhibit significant divergence across different dimensions. In \cref{fig:cross_dim}, Seed-1.6-V. (\#2) ranks only \#7 in the \textit{Dis. Txt.} task, suggesting a potential weakness in OCR capabilities, yet it achieves \#1 in all \textit{Spatial} tasks. Such discrepancies reveal potential imbalances in training data, providing a clear direction for domain-specific enhancements.

\subsection{Model Family Behaviors.} 
\begin{figure}[h]
  \centering
   \includegraphics[width=0.9\linewidth]{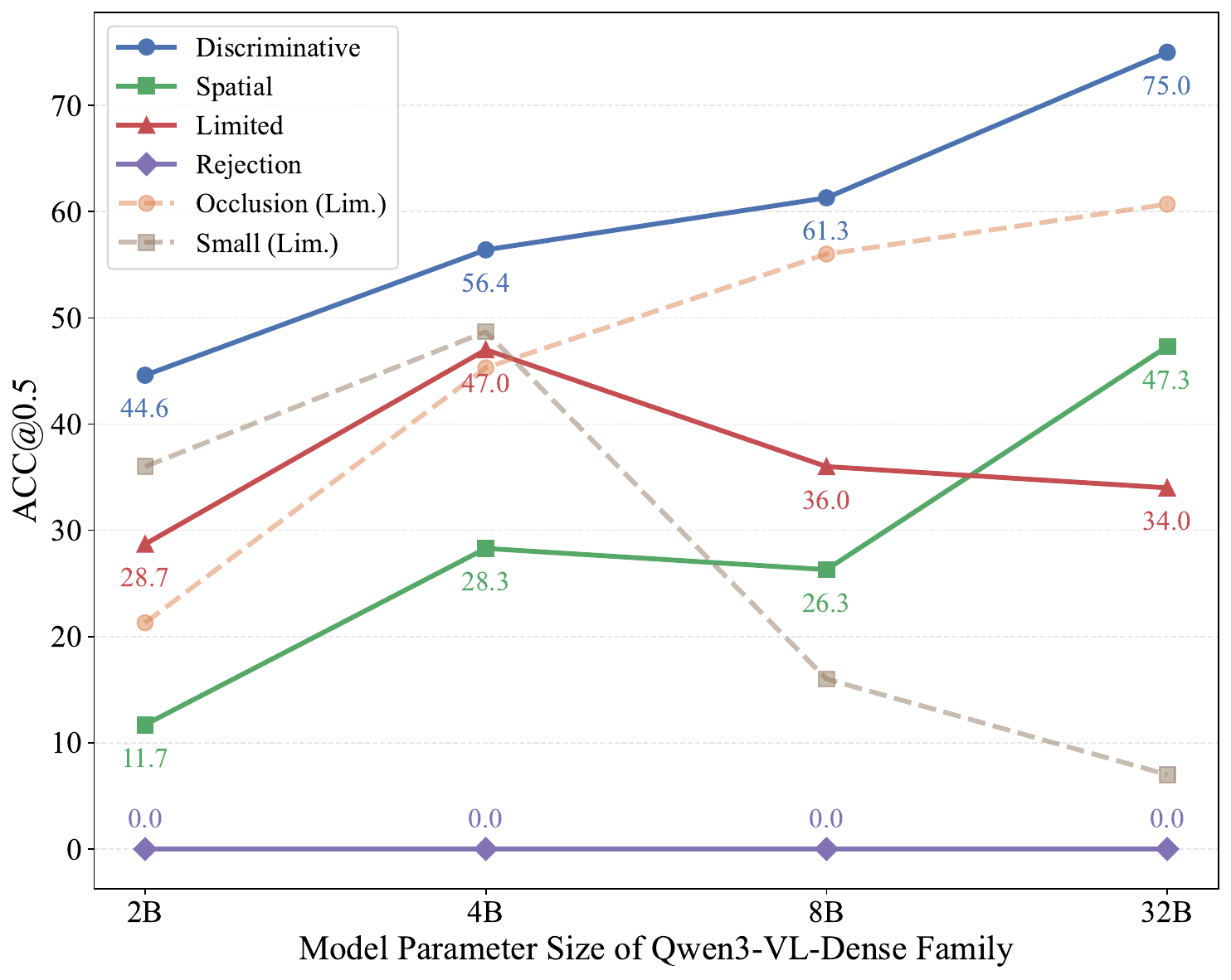}
   \caption{Qwen3-VL Family Behavior.}
   \label{fig:qwen3_family}
\end{figure}

\cref{fig:qwen3_family} explores the behaviors of models within the Qwen3-VL dense model family (2B-32B). We observe: (1) Models of all size score 0 on \textit{Rejection}, demonstrating high model-family consistency. (2) Performance on \textit{Discriminative} and \textit{Spatial} correlates positively with model scaling, while no obvious trend on \textit{Limited}. (3) Subtask decomposition reveals the anomaly of \textit{Limited} stems from \textit{Small}, providing insight for further analysis.

\section{Evaluation Prompts}

\subsection{Prompt Templates}
\cref{tab:eval_prompt} presents the unified prompt template employed for all evaluations conducted on \DATASET{}.
\cref{tab:tts_mllm_prompt} details the prompt template used for the MLLM judge baseline during the best-of-N selection in our test-time scaling analysis.
\cref{tab:tts_text_prompt} shows the prompt template used for the text-only LLM judge to select the optimal response based on thinking trajectory during our test-time scaling analysis.

\subsection{Prompt Robustness}
We use 5 distinct prompts to evaluate Qwen3-VL-8B and Llama-Nemotron and yield accuracies of 31.36$\pm$2.02 and 10.07$\pm$0.44, matching previous scores and showing little variance. We also find 98.61\% of Qwen3-VL-8B responses follow the instructed format, and fix the remaining by comprehensive rules to minimize instruction-following errors.

\section{Human Rejection Verification}

Given the poor performance of models on the Rejection category, we conduct a human verification study to validate the correctness and quality of our data. We randomly sample 100 instances from \DATASET{} for human binary classification (Reject/Non-Reject) annotation. To mitigate the risk of human annotators taking linguistic shortcuts based on distinct description styles, we restrict the sampling to the Discriminative and Rejection categories only, as their referring expressions exhibit structural similarity. The final sampled set includes 51 instances from the Rejection category. Considering Rejection as the positive class, human annotators achieve an Accuracy of 91\%, with a Precision of 88.24\%, a Recall of 93.75\%, and an F1-score of 90.91\%.

\section{Commercial Model Notes}

Regarding the evaluation of commercial models, we make a specific adjustment for the Gemini-2.5 series: we modify the required coordinate format in the prompt template (\cref{tab:eval_prompt}) to [y1, x1, y2, x2]. This modification is implemented because we observe that Gemini-2.5 is significantly more receptive to this output format, resulting in a measurable improvement in accuracy.

We do not report the evaluation results for GPT-5, Claude-Sonnet-4.5, and Grok-4 due to issues with their output. From cases in \cref{tab:commercial_case}, we observe that the coordinates produced by these models using the unified prompt template (\cref{tab:eval_prompt}) suffered from substantial displacement and distortion, regardless of whether the output is interpreted as absolute pixel coordinates (red bounding box) or 0-999 normalized relative coordinates (blue bounding box). Furthermore, we fail to find an alternative coordinate format that yields usable results for these models.

\section{Tool Use Results}

We also conduct an evaluation of Claude-Sonnet-4.5 utilizing PyVision~\cite{zhao2025pyvision} for tool use. The total accuracy is 12.4\%. The detailed subtask breakdown is as follows: Discriminative: 19.1\% (App.: 15.4\%, Cmp.: 32\%, Txt.: 10\%, Sta.: 19.2\%); Spatial: 13.3\% (Rel.: 13.3\%, Cnt.: 13.3\%); Limited: 9.0\% (Occ.: 10.7\%, Sml.: 7.3\%); and Rejection: 9.5\% (App.: 10\%, Cmp.: 7.8\%, Txt.: 14\%, Sta.: 6\%).

We assume that the model's ability to utilize tool use for multi-step cropping and magnification of 8K image to localize tiny objects should yield improved performance on the Limited\_Small subcategory. However, the observed accuracy (7.3\%) falls below our expectations. Through case study in \cref{tab:tool_case}, we find that subtle offset of bounding box size and position is a significant contributing factor to this unsatisfactory result.

\section{Examples for Each L-2 Subcategory}

In \cref{tab:dis_app} through \cref{tab:rej_sta}, we provide precise task definitions for all twelve L-2 subcategories in \DATASET{}, and present one representative example for each. In all displayed examples, the red bounding box indicates the correct ground-truth object.

\clearpage

\input{appendix/iou}

\input{appendix/eval_prompt}

\input{appendix/mllm_judge_prompt}

\input{appendix/text_judge_prompt}

\input{appendix/commercial_case}

\input{appendix/tool_case}

\input{appendix/examples}

%% file: appendix/iou.tex
\begin{table*}
\small 
  \centering
  \setlength{\tabcolsep}{3pt}
  \caption{Evaluation results across different IoU thresholds on \DATASET{}. All settings and abbreviations are the same as in \S4.}
  \label{tab:iou}
  
  \begin{tabular}{l|rrr|rrr|rrr|rrr}
    \toprule
     \multicolumn{1}{c}{\multirow{2}{*}{\textbf{Model}}} & \multicolumn{3}{c}{\textbf{Discriminative}} & \multicolumn{3}{c}{\textbf{Spatial}} & \multicolumn{3}{c}{\textbf{Limited}} & \multicolumn{3}{c}{\textbf{Total}} \\
    \cmidrule(lr){2-4} \cmidrule(lr){5-7} \cmidrule(lr){8-10} \cmidrule(lr){11-13}
     \multicolumn{1}{c}{} & Acc$_{0.75}$ & Acc$_{0.9}$ & mAcc & Acc$_{0.75}$ & Acc$_{0.9}$ & mAcc & Acc$_{0.75}$ & Acc$_{0.9}$ & mAcc & Acc$_{0.75}$ & Acc$_{0.9}$ & mAcc \\
    \midrule
    Phi-4-Multimodal & 0.0 & 0.0 & 0.1 & 0.0 & 0.0 & 0.2 & 0.0 & 0.0 & 0.0 & 0.0 & 0.0 & 0.1 \\
    Llama-4-Maverick & 8.8 & 0.5 & 9.5 & 8.3 & 0.7 & 9.8 & 1.7 & 0.0 & 1.9 & 6.0 & 1.5 & 6.6 \\
    Llama-4-Scout & 7.4 & 0.5 & 8.0 & 3.7 & 0.0 & 4.9 & 1.3 & 0.0 & 1.4 & 3.5 & 0.6 & 4.0 \\
    LLaVA-O.V.-1.5-8B & 2.9 & 0.5 & 4.0 & 0.7 & 0.0 & 1.1 & 0.0 & 0.0 & 0.7 & 0.8 & 0.1 & 1.4 \\
    Mistral-3.2-24B & 0.0 & 0.0 & 1.1 & 0.0 & 0.0 & 0.8 & 0.0 & 0.0 & 0.0 & 0.0 & 0.0 & 0.5 \\
    Gemma-3-27B & 0.0 & 0.0 & 0.4 & 0.0 & 0.0 & 0.0 & 0.0 & 0.0 & 0.0 & 0.0 & 0.0 & 0.1 \\
    Llama-Nemotron-8B & 14.7 & 3.9 & 14.9 & 2.7 & 0.7 & 2.9 & 4.7 & 2.0 & 5.1 & 6.3 & 2.7 & 6.5 \\
    MiniCPM-V-4.5 & 1.0 & 0.0 & 2.4 & 0.7 & 0.0 & 1.4 & 0.0 & 0.0 & 1.1 & 0.4 & 0.0 & 1.2 \\
    InternVL3.5-8B & 2.0 & 1.5 & 3.2 & 2.3 & 0.0 & 2.2 & 0.7 & 0.3 & 0.9 & 1.6 & 0.7 & 1.9 \\
    InternVL3.5-A28B & 16.2 & 6.4 & 16.4 & 14.3 & 3.0 & 13.9 & 6.7 & 3.3 & 7.6 & 9.6 & 3.2 & 9.8 \\
    Keye-VL-1.5-8B & 8.3 & 0.5 & 9.8 & 1.3 & 0.0 & 2.6 & 0.7 & 0.3 & 1.6 & 2.3 & 0.2 & 3.2 \\
    MiMo-VL-7B-RL & 33.8 & 13.2 & 30.6 & 14.0 & 5.0 & 12.8 & 8.0 & 3.0 & 8.1 & 13.4 & 5.1 & 12.5 \\
    GLM-4.5V & 44.1 & 32.4 & 42.5 & 37.7 & 24.7 & 34.5 & 19.0 & 8.3 & 18.2 & 26.0 & 16.5 & 24.5 \\
    Qwen2.5-VL-7B & 24.0 & 12.3 & 22.4 & 10.3 & 4.0 & 9.3 & 9.3 & 1.3 & 8.4 & 10.8 & 4.2 & 9.9 \\
    Qwen2.5-VL-32B & 38.7 & 15.2 & 33.6 & 29.0 & 13.0 & 27.2 & 9.7 & 3.0 & 10.8 & 19.4 & 7.9 & 18.2 \\
    Qwen2.5-VL-72B & 42.2 & 21.1 & 37.4 & 30.0 & 14.7 & 28.4 & 14.3 & 4.3 & 14.4 & 22.4 & 10.5 & 21.0 \\
    Qwen3-VL-2B & 39.2 & 31.9 & 37.8 & 10.3 & 8.0 & 9.9 & 15.3 & 7.0 & 16.5 & 15.6 & 10.9 & 15.6 \\
    Qwen3-VL-4B & 52.9 & 43.1 & 50.2 & 25.7 & 20.0 & 24.3 & 29.7 & 11.7 & 29.4 & 27.3 & 18.2 & 26.2 \\
    Qwen3-VL-8B & 57.4 & 47.1 & 55.0 & 23.7 & 21.0 & 23.3 & 27.3 & 16.7 & 26.5 & 26.9 & 20.8 & 26.0 \\
    Qwen3-VL-32B & 71.1 & 54.9 & 65.9 & 44.0 & 32.7 & 41.3 & 28.3 & 19.0 & 26.3 & 36.0 & 26.6 & 33.6 \\
    Qwen3-VL-A3B & 61.3 & 50.5 & 57.8 & 27.3 & 20.0 & 25.6 & 32.0 & 15.3 & 31.4 & 30.1 & 20.8 & 28.7 \\
    Qwen3-VL-A22B & 66.7 & 54.9 & 63.3 & 46.3 & 38.3 & 44.1 & 37.3 & 21.7 & 36.7 & 38.5 & 29.1 & 37.0 \\
    \midrule
    Gemini-2.5-Pro & 22.5 & 11.8 & 22.9 & 23.0 & 12.3 & 21.7 & 2.7 & 0.7 & 3.5 & 13.6 & 7.7 & 13.6 \\
    Gemini-2.5-Flash & 27.5 & 14.7 & 26.6 & 19.0 & 13.3 & 19.0 & 6.3 & 1.0 & 7.2 & 13.1 & 7.3 & 13.2 \\
    Seed-1.6-V.-250815 & 55.4 & 41.7 & 51.7 & 51.3 & 35.7 & 48.0 & 30.7 & 19.0 & 30.1 & 35.9 & 25.0 & 34.0 \\
    \bottomrule
    
  \end{tabular}
\end{table*}

%% file: appendix/eval_prompt.tex
\begin{table}[t]
\small
\centering
\scalebox{1}{
\begin{tcolorbox}[colback=gray!00,%
                  colframe=black,%
                  width=\linewidth,%
                  arc=1.5mm, auto outer arc,
                  breakable,
                  left=0.9mm, right=0.9mm,
                  boxrule=0.9pt,
                 ]

\textless image\textgreater \\

All spatial relationships are defined from the viewer's perspective, where 'front' means closer to the viewer and 'back' means farther from the viewer. Please provide the bounding box coordinate of the object the following statement describes: \\

\{description\} \\

Ensure that all details mentioned about the object are accurate. Provide at most one bounding box. If a matching object is found, provide its bounding box as a JSON in the format \{``bbox\_2d": [x1, y1, x2, y2]\}. If no matching object is found, output \{``bbox\_2d": null\}.

\end{tcolorbox}}
\caption{Prompt template for all evaluations on \DATASET{}.}
\label{tab:eval_prompt}
\end{table}

%% file: appendix/mllm_judge_prompt.tex
\begin{table}[t]
\small
\centering
\scalebox{1}{
\begin{tcolorbox}[colback=gray!00,%
                  colframe=black,%
                  width=\linewidth,%
                  arc=1.5mm, auto outer arc,
                  breakable,
                  left=0.9mm, right=0.9mm,
                  boxrule=0.9pt,
                 ]

\textless image\textgreater \\

\# Role and Task \\

You are an expert-level Visual Grounding Adjudicator. Your task is to evaluate two proposed bounding boxes (Bbox A and Bbox B) for a given image and user instruction, and determine which one is the more accurate and superior choice. \\

\# Input \\

\textbf{Instruction}: \{instruction\} \\
\textbf{Bbox A}: \{bbox\_a\} \\
\textbf{Bbox B}: \{bbox\_b\} \\

\# Output

Explain your reasoning, then conclude with your final choice in the format \textbackslash boxed\{A\} or \textbackslash boxed\{B\}.

\end{tcolorbox}}
\caption{Prompt template for multimodal judge models in test-time scaling analysis.}
\label{tab:tts_mllm_prompt}
\end{table}

%% file: appendix/text_judge_prompt.tex
\begin{table}[t]
\small
\centering
\scalebox{1}{
\begin{tcolorbox}[colback=gray!00,%
                  colframe=black,%
                  width=\linewidth,%
                  arc=1.5mm, auto outer arc,
                  breakable,
                  left=0.9mm, right=0.9mm,
                  boxrule=0.9pt,
                 ]

\# Role and Task \\

You are a rigorous AI reasoning process analyst. Your task is to compare the two responses provided (Response A and Response B) based on the five principles below and select the superior one. \\

\# Core Evaluation Principles \\

All of your judgments MUST be strictly based on the following five points: \\

1. \textbf{Instruction Understanding}: Evaluate whether the model has correctly and comprehensively understood the description in the user's instruction, including all details, constraints, and limitations. \\
2. \textbf{Visual Observation}: Evaluate whether the model has comprehensively and meticulously observed the image, identifying as many objects and their attributes or spatial relationships as possible. \\
3. \textbf{Logical Reasoning}: Evaluate whether each step of the reasoning is logical and free of contradictions, fallacies, or unsubstantiated leaps. \\ 
4. \textbf{Analytical Rigor}: Evaluate whether the conclusion was reached hastily or formed after carefully analyzing and comparing multiple possibilities. \\
5. \textbf{Conclusion Support}: Evaluate whether the final answer is strongly supported and uniquely derived from the thought process, rather than being disconnected from it. \\

\# Input \\

[Original Task Instruction] \\
\{instruction\} \\

[Full Content of Response A] \\
\{response\_a\} \\

[Full Content of Response B] \\
\{response\_b\} \\

\# Output \\

Your final selection must be \textbf{only} one of the following two lines, with no other text before or after: \textbackslash boxed\{A\} or \textbackslash boxed\{B\}.

\end{tcolorbox}}
\caption{Prompt template for text-only judge models in test-time scaling analysis.}
\label{tab:tts_text_prompt}
\end{table}

%% file: appendix/commercial_case.tex
\begin{table}[t]
\small
\centering
\scalebox{1}{

\begin{tcolorbox}[colback=gray!00,%
                  colframe=black,%
                  width=8.5cm,%
                  arc=1.5mm, auto outer arc,
                  breakable,
                  left=0.9mm, right=0.9mm,
                  boxrule=0.9pt, colbacktitle = black!65!black
                 ]

\textbf{Description:}  \\
This is a small, light green plastic stool with a top and four tapered legs that splay slightly outwards. The top surface has a subtle pattern. A small sticker is attached to it. Its size suggests it's a common, lightweight outdoor seating option. \\
\textbf{Correct Answer:} [544, 1102, 940, 1498]
\begin{center}
\includegraphics[width = 0.7\linewidth]{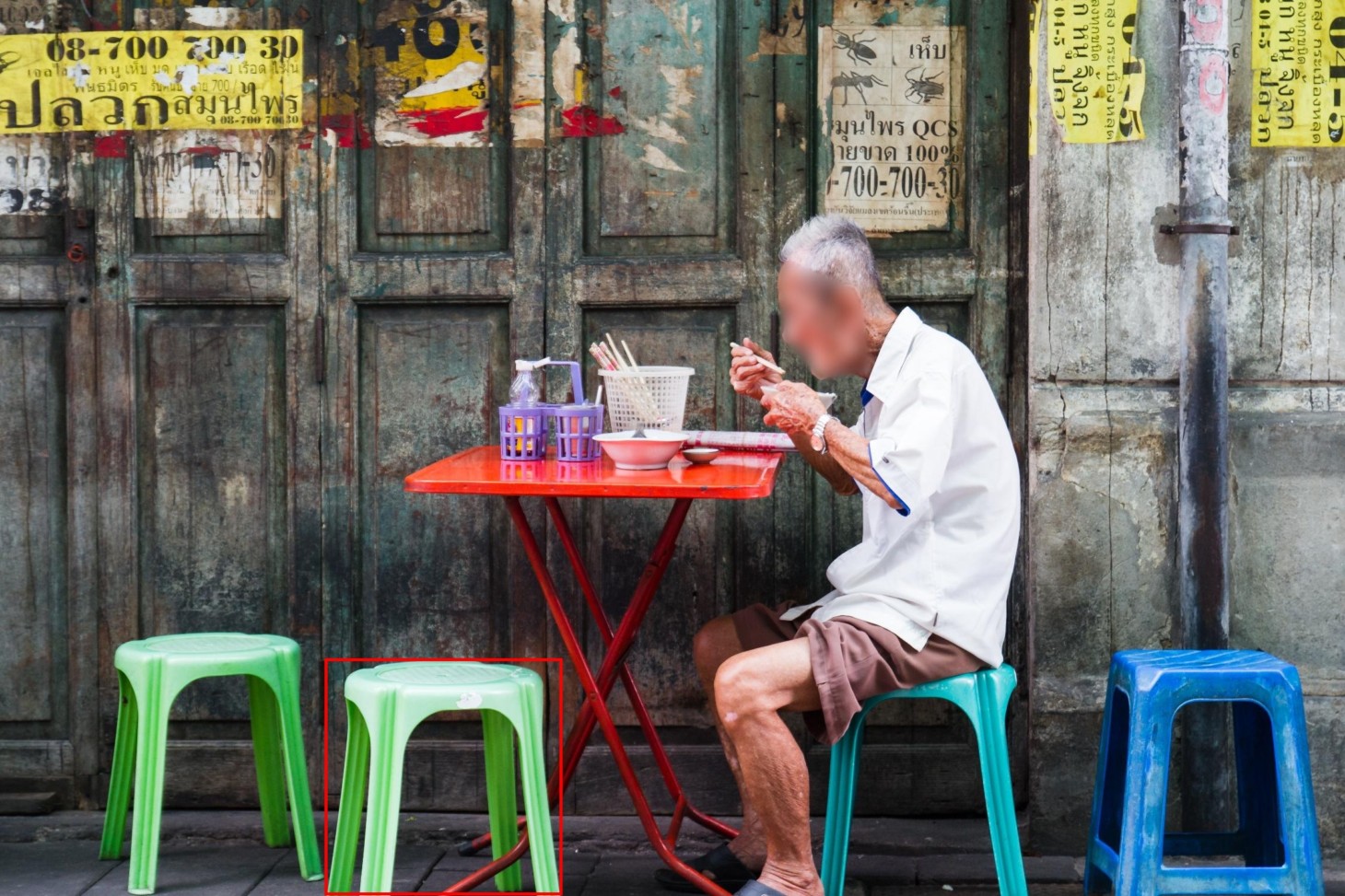}
\end{center}
\textbf{GPT-5 Answer:} [320, 600, 520, 760]
\begin{center}
\includegraphics[width = 0.7\linewidth]{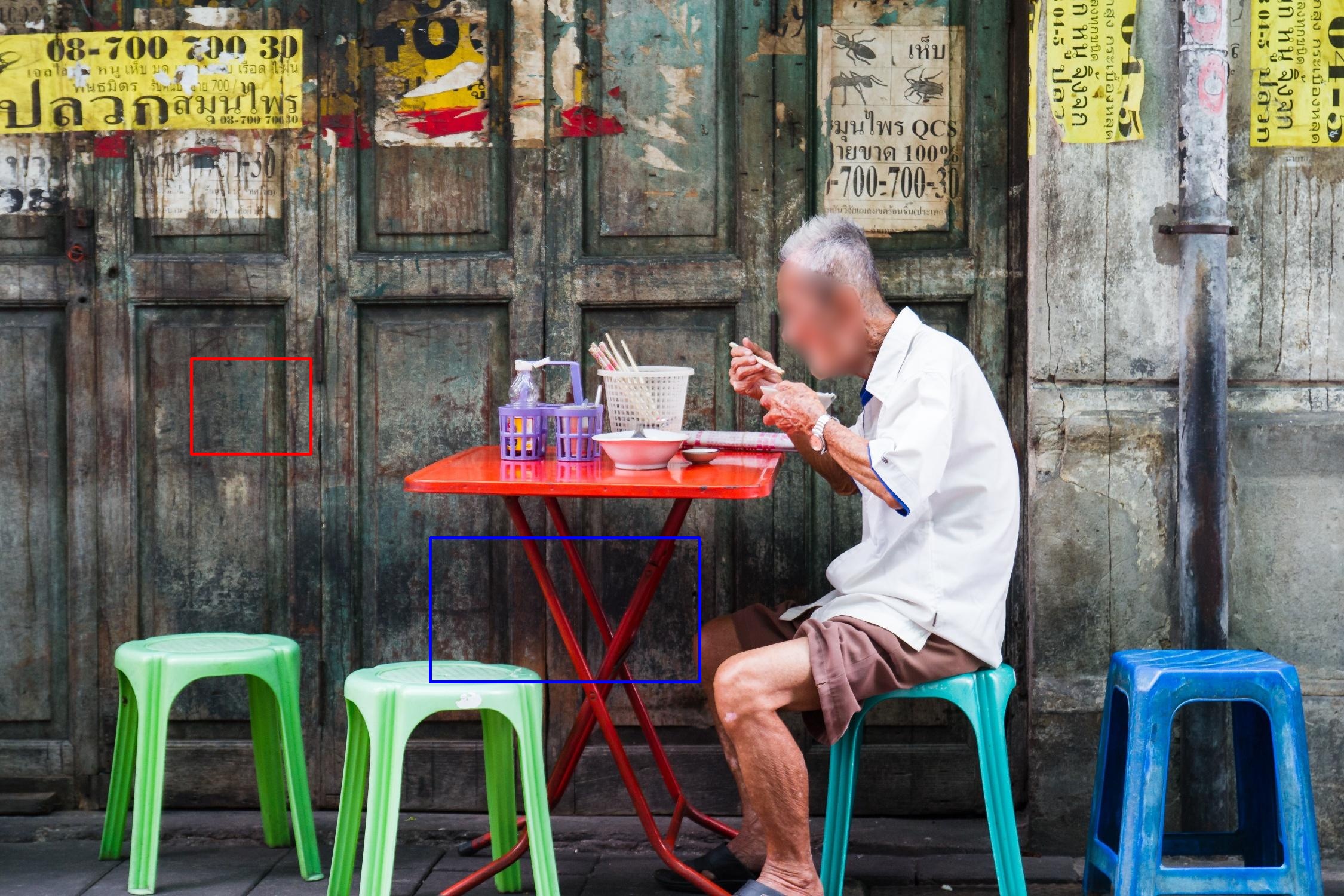}
\end{center}
\textbf{Claude-4.5 Answer:} [89, 632, 301, 869]
\begin{center}
\includegraphics[width = 0.7\linewidth]{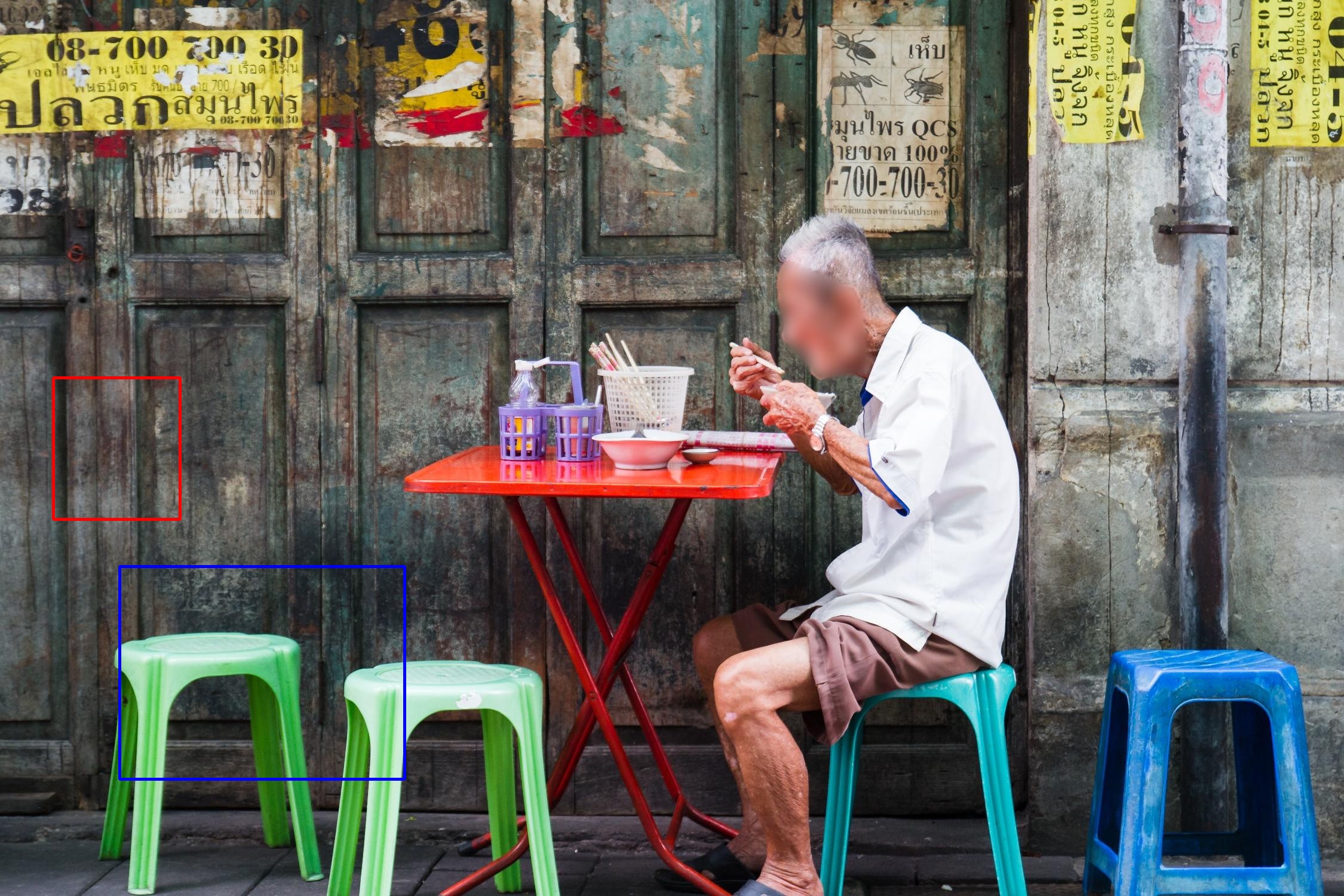}
\end{center}
\textbf{Grok-4 Answer:} [59, 322, 130, 410]
\begin{center}
\includegraphics[width = 0.7\linewidth]{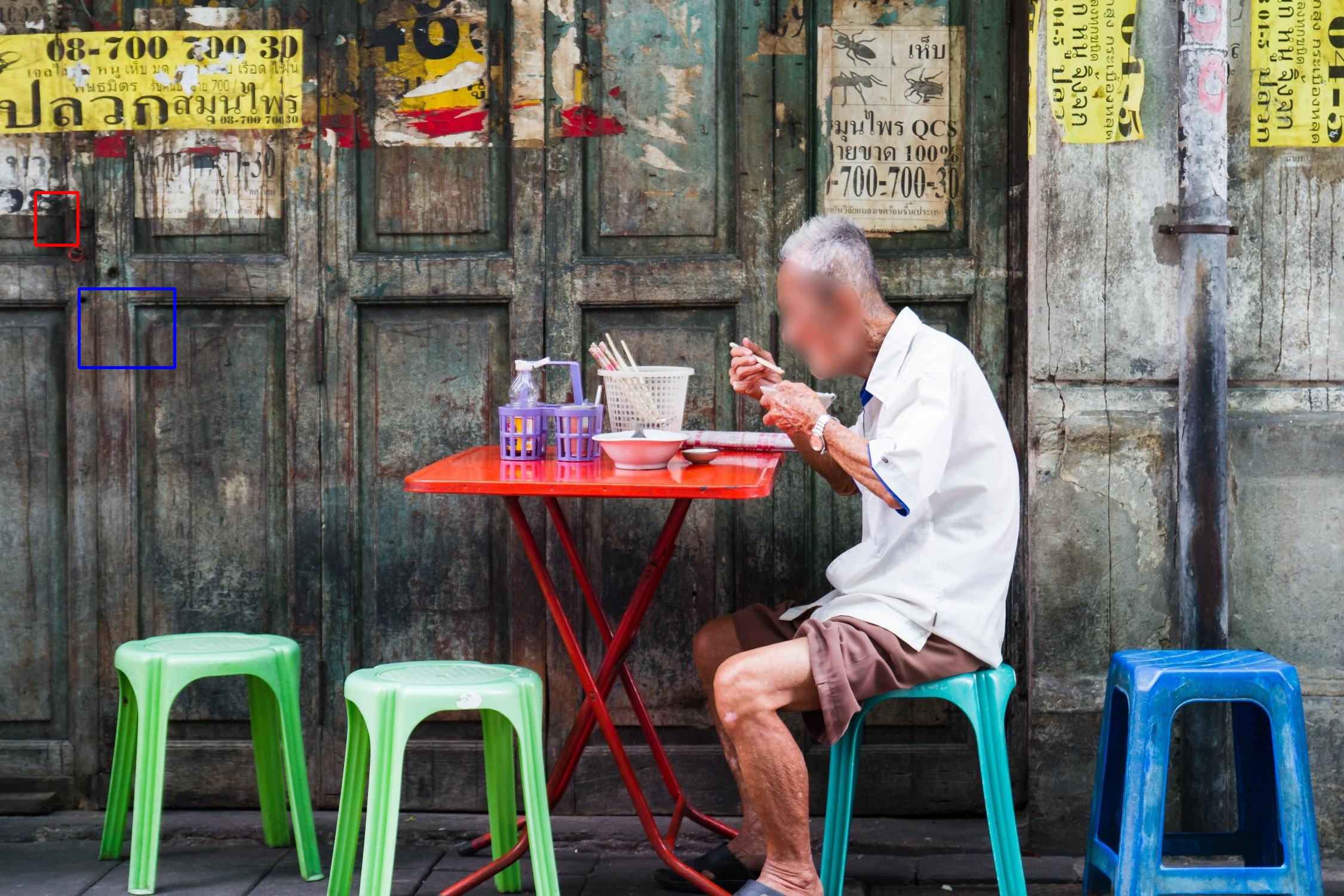}
\end{center}

\end{tcolorbox}
}
\caption{Cases of outputs from unreported commercial models.}
\label{tab:commercial_case}
\end{table}

%% file: appendix/tool_case.tex
\begin{table}[t]
\small
\centering
\scalebox{1}{

\begin{tcolorbox}[colback=gray!00,%
                  colframe=black,%
                  width=8.5cm,%
                  arc=1.5mm, auto outer arc,
                  breakable,
                  left=0.9mm, right=0.9mm,
                  boxrule=0.9pt, colbacktitle = black!65!black
                 ]

\textbf{Description:}  \\
The object is a young girl. She is squatting on a wooden raft or platform by the water. \\

\textbf{Original Image:} width=7680, height=5046
\begin{center}
\includegraphics[width = 0.7\linewidth]{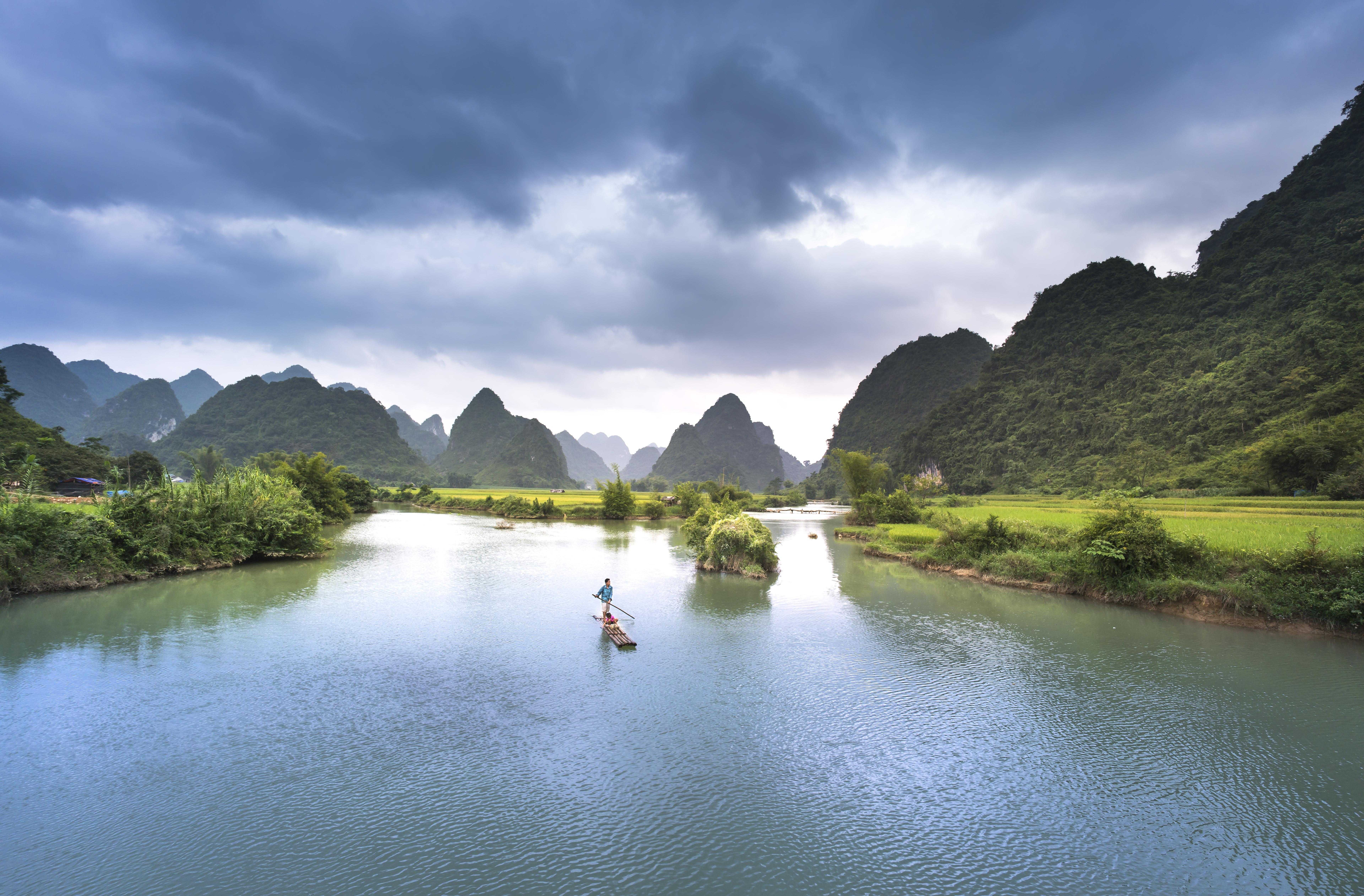}
\end{center}
\textbf{Correct Answer:} [3389, 3448, 3487, 3530]
\begin{center}
\includegraphics[width = 0.7\linewidth]{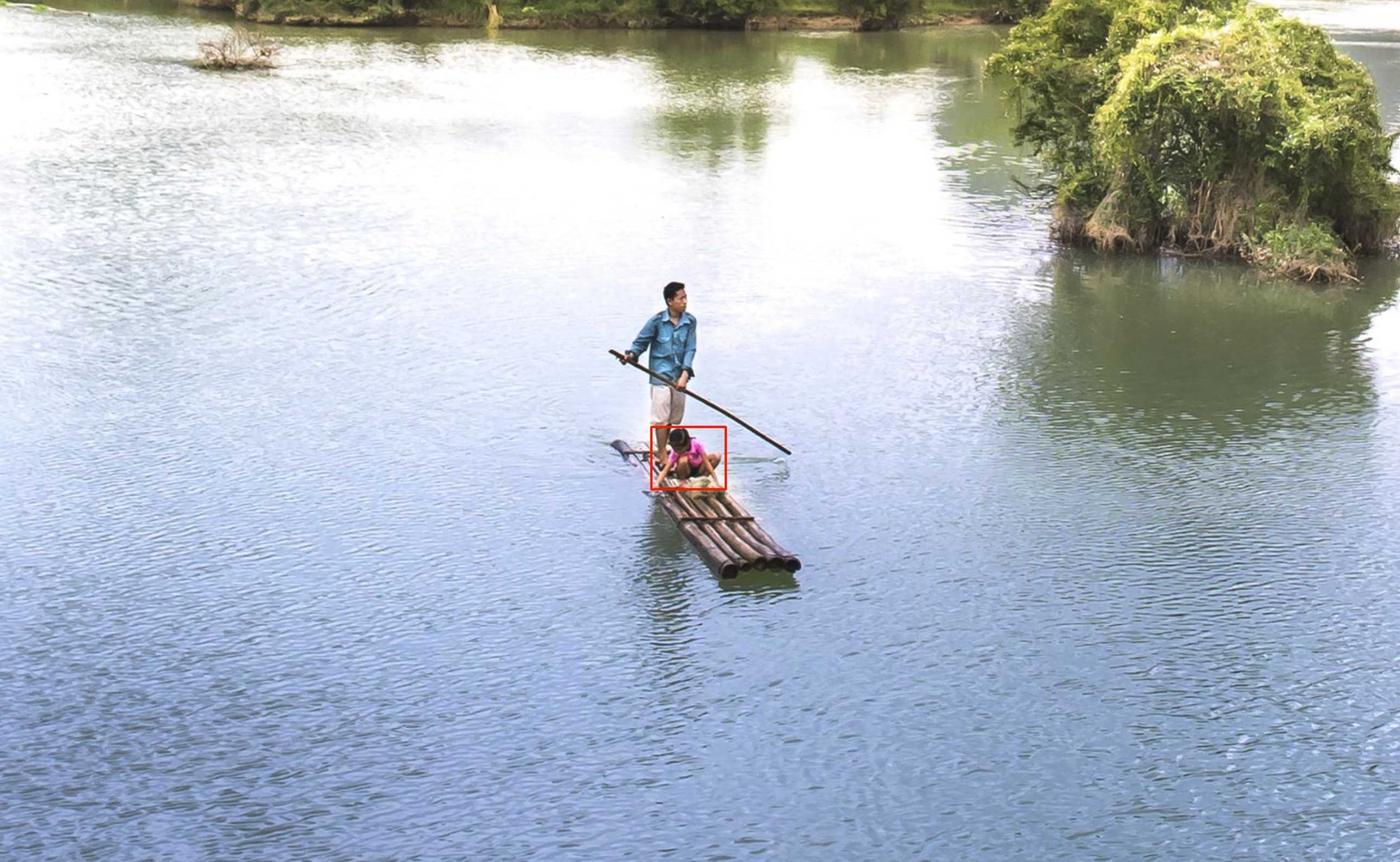}
\end{center}
\textbf{Claude-4.5 Answer:} [0.4323, 0.6877, 0.4544, 0.7134]
\begin{center}
\includegraphics[width = 0.7\linewidth]{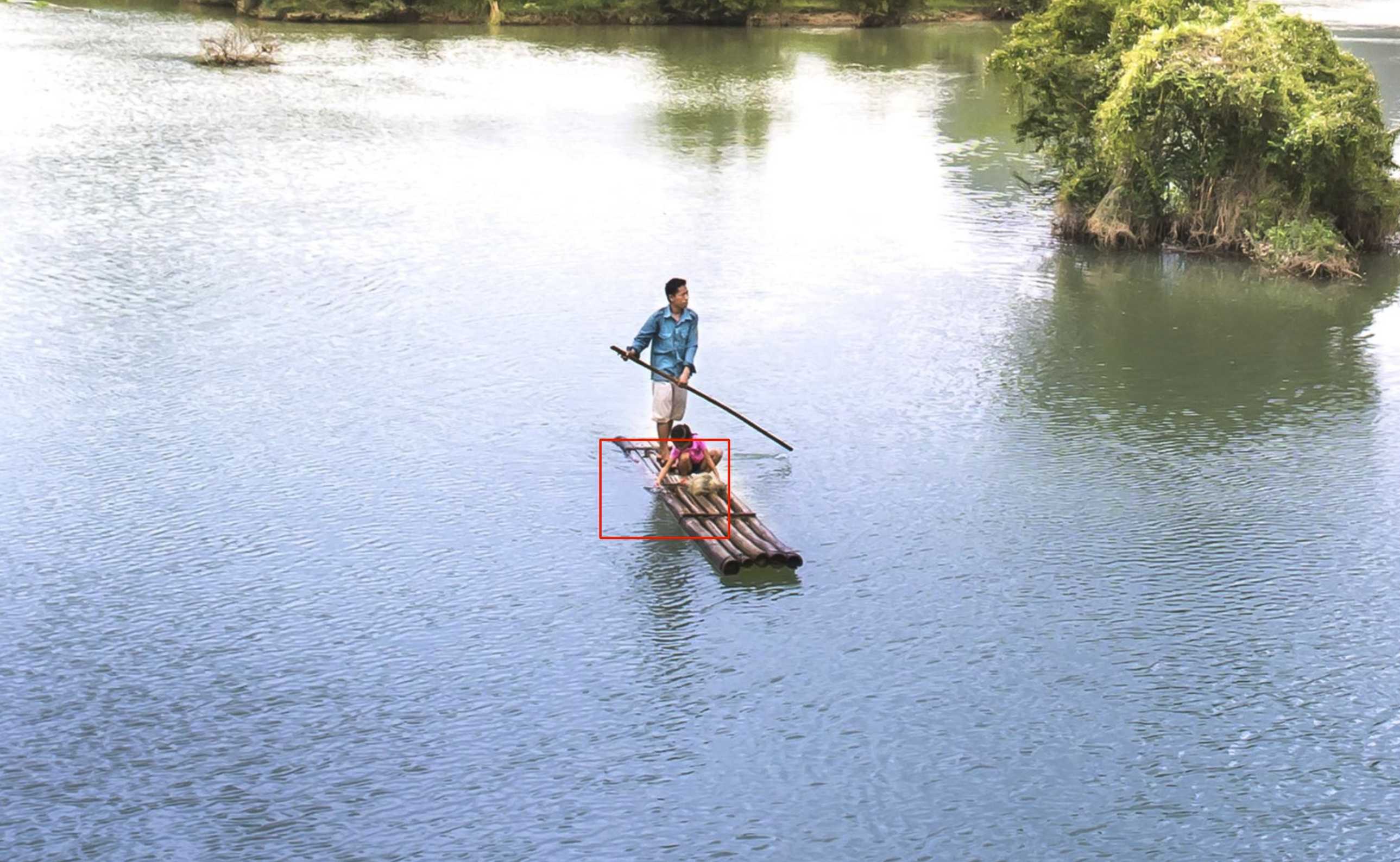}
\end{center}

\end{tcolorbox}
}
\caption{Case of Claude-Sonnet-4.5 output with tool use.}
\label{tab:tool_case}
\end{table}

%% file: appendix/examples.tex
\begin{table}[t]
\small
\centering
\scalebox{1}{

\begin{tcolorbox}[colback=gray!00,%
                  colframe=black,%
                  width=8.5cm,%
                  arc=1.5mm, auto outer arc,
                  breakable,
                  left=0.9mm, right=0.9mm,
                  boxrule=0.9pt, colbacktitle = black!65!black,
                  title = {Subcategory 1: Discriminative\_Appearance}
                 ]
{\textbf{Definition:} Distinguishing objects based on subtle visual attributes like color or texture.} \\
\begin{center}
\includegraphics[width = 0.8\linewidth]{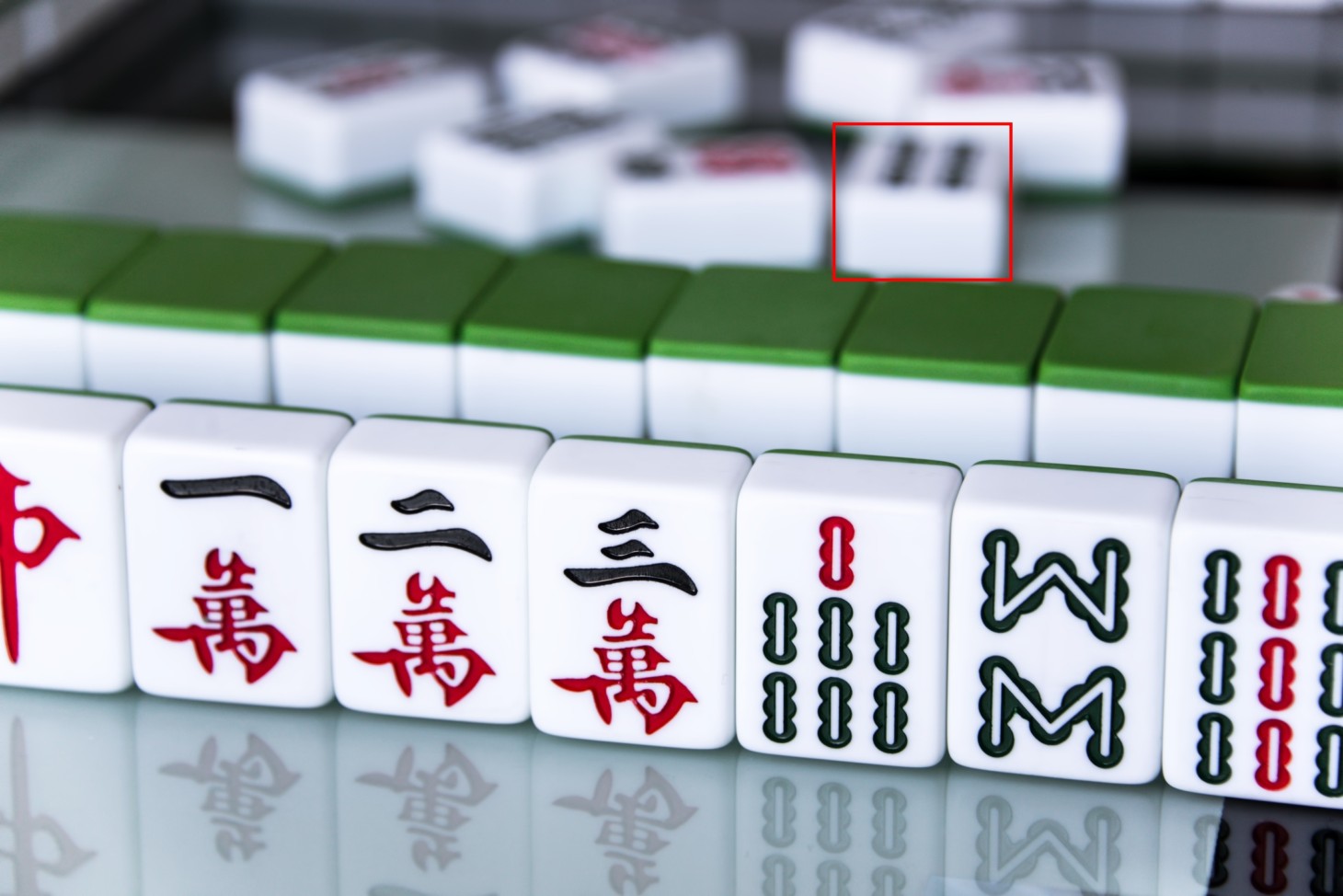}
\end{center}
\textbf{Description:} 
\\This is a white cube, likely a Mahjong tile, with a smooth, reflective surface. \textbf{On its top face, there are two blurry, dark vertical markings, which appear to be thin lines or abstract shapes, rendered out of focus.} The material seems to be a hard, glossy substance like plastic or ceramic.
\\
\end{tcolorbox}
}
\caption{An example of Discriminative\_Appearance Subtask.}
\label{tab:dis_app}
\end{table}

\begin{table}[t]
\small
\centering
\scalebox{1}{

\begin{tcolorbox}[colback=gray!00,%
                  colframe=black,%
                  width=8.5cm,%
                  arc=1.5mm, auto outer arc,
                  breakable,
                  left=0.9mm, right=0.9mm,
                  boxrule=0.9pt, colbacktitle = black!65!black,
                  title = {Subcategory 2: Discriminative\_Component}
                 ]
{\textbf{Definition:} Distinguishing targets based on the presence or absence of a specific structural component.} \\
\begin{center}
\includegraphics[width = 0.8\linewidth]{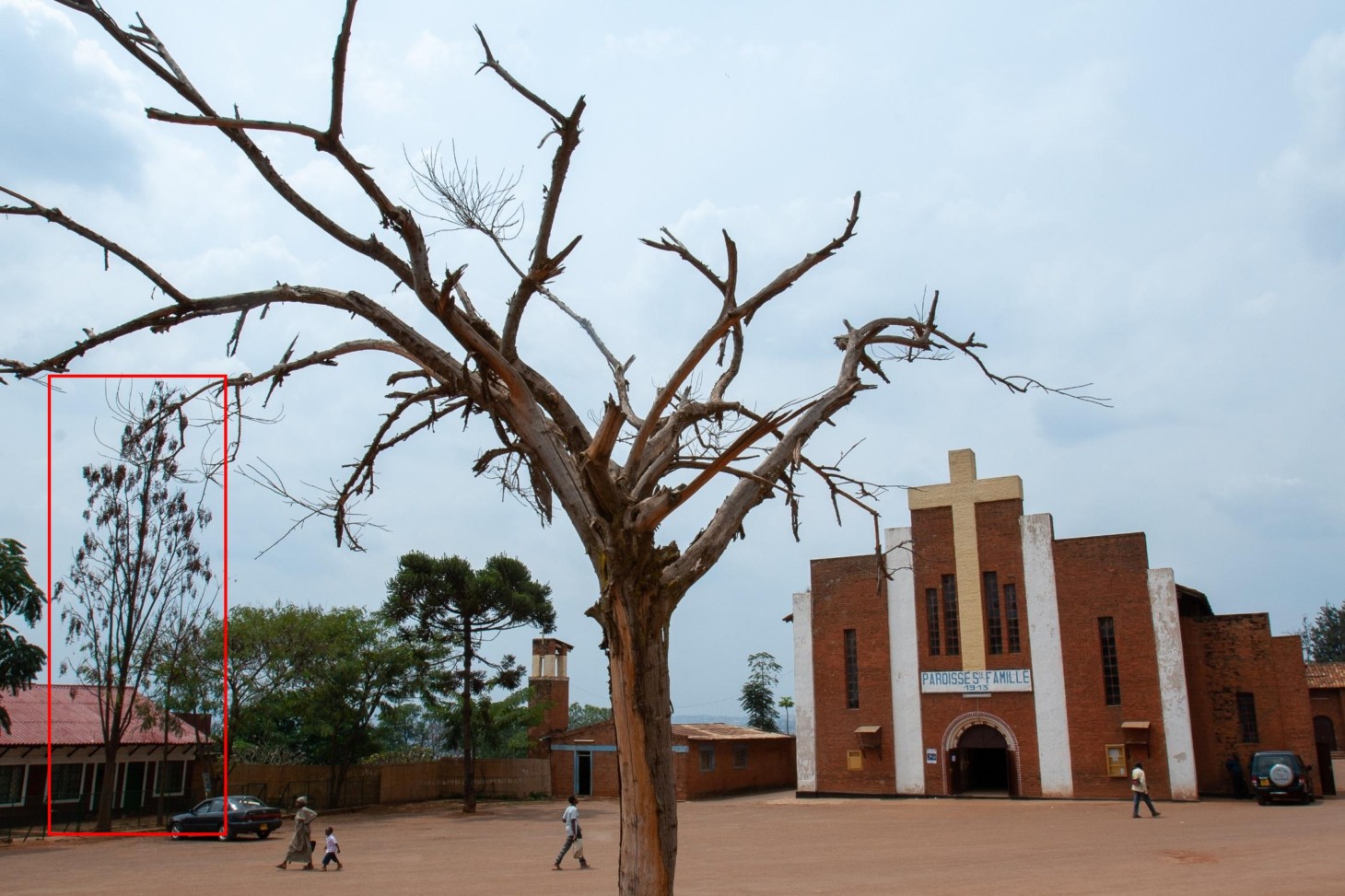}
\end{center}
\textbf{Description:} 
\\This is a tall, slender tree with a relatively straight, thin trunk and sparse, upright branches. The trunk and branches are primarily dark brown to reddish-brown, suggesting sparse foliage. The texture appears rough and natural. \textbf{There are a few brown leaves concentrated on some of the upper branches}, indicating it might be a deciduous tree in a dry season or a tree with naturally sparse foliage.
\\
\end{tcolorbox}
}
\caption{An example of Discriminative\_Component Subtask.}
\label{tab:dis_cmp}
\end{table}

\begin{table}[t]
\small
\centering
\scalebox{1}{

\begin{tcolorbox}[colback=gray!00,%
                  colframe=black,%
                  width=8.5cm,%
                  arc=1.5mm, auto outer arc,
                  breakable,
                  left=0.9mm, right=0.9mm,
                  boxrule=0.9pt, colbacktitle = black!65!black,
                  title = {Subcategory 3: Discriminative\_Text}
                 ]
{\textbf{Definition:} Distinguishing targets based on textual information embedded within the image.} \\
\begin{center}
\includegraphics[width = 0.8\linewidth]{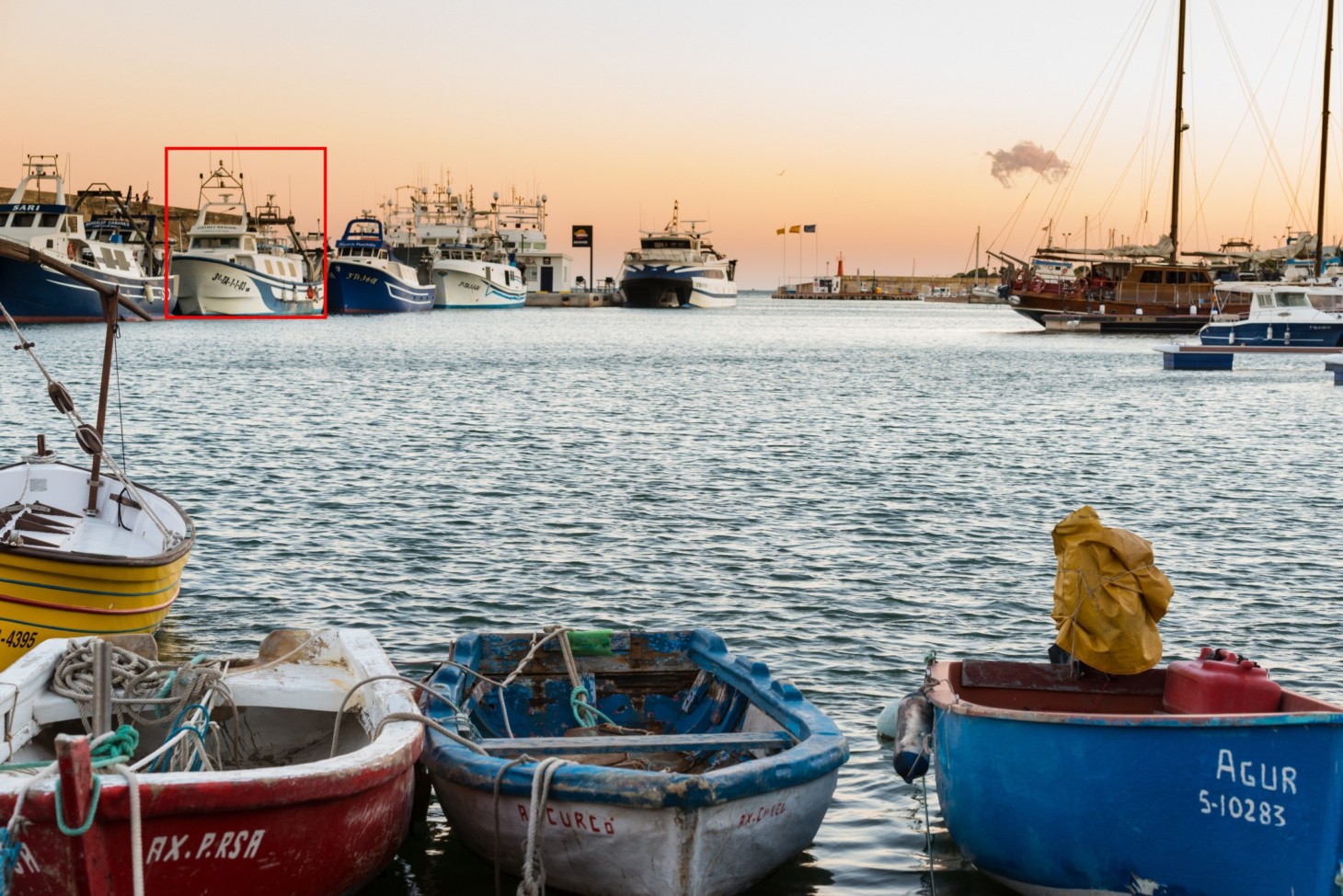}
\end{center}
\textbf{Description:} 
\\This object is a boat. T\textbf{he number '1-1-03' is visible on its hull.}
\\
\end{tcolorbox}
}
\caption{An example of Discriminative\_Text Subtask.}
\label{tab:dis_txt}
\end{table}

\begin{table}[t]
\small
\centering
\scalebox{1}{

\begin{tcolorbox}[colback=gray!00,%
                  colframe=black,%
                  width=8.5cm,%
                  arc=1.5mm, auto outer arc,
                  breakable,
                  left=0.9mm, right=0.9mm,
                  boxrule=0.9pt, colbacktitle = black!65!black,
                  title = {Subcategory 4: Discriminative\_State}
                 ]
{\textbf{Definition:} Distinguishing objects based on their dynamic or static condition.} \\
\begin{center}
\includegraphics[width = 0.8\linewidth]{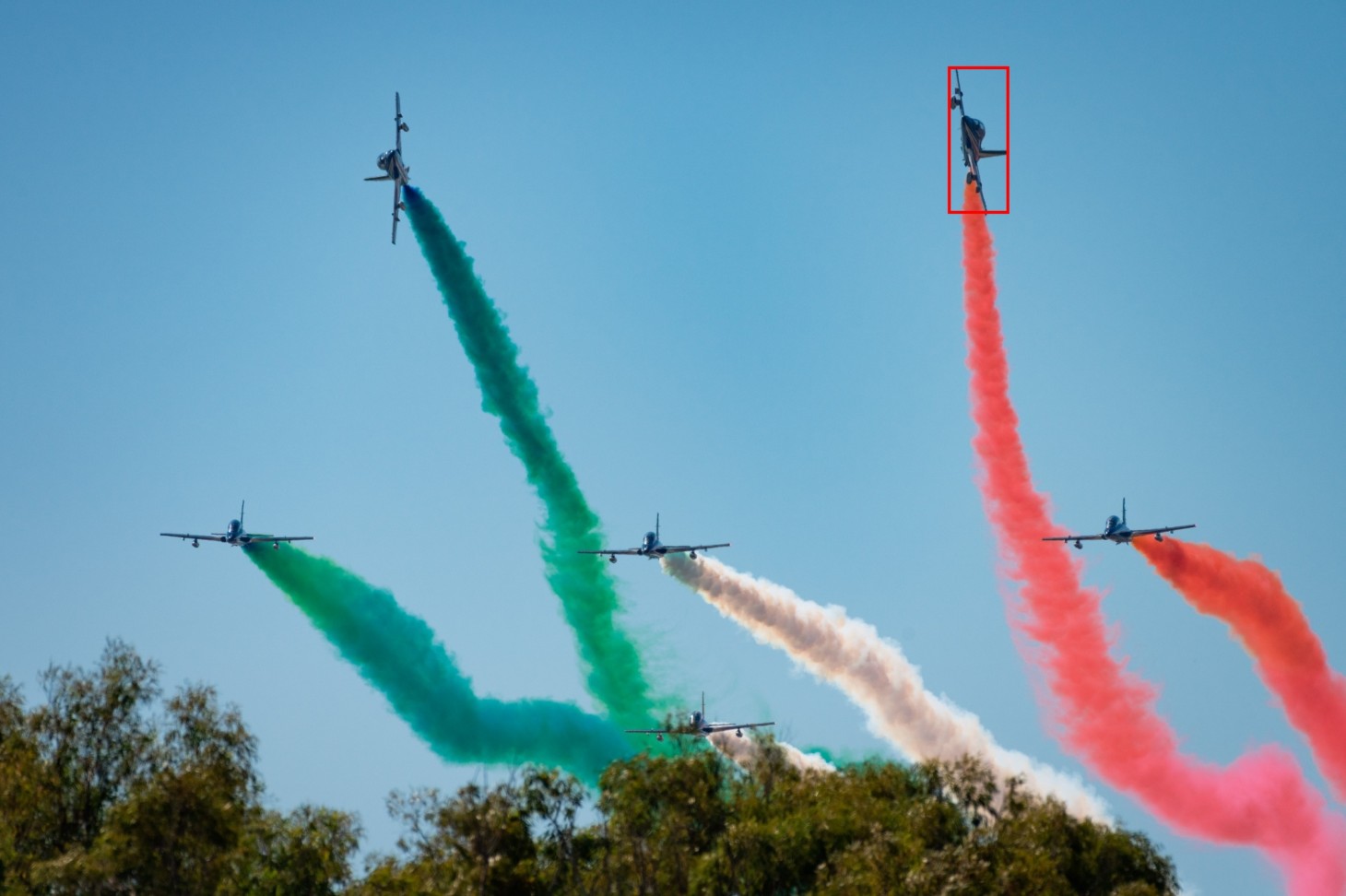}
\end{center}
\textbf{Description:} 
\\The jet is dark-colored, appearing black or very dark navy, with a sleek, aerodynamic design characteristic of a military training aircraft. \textbf{The jet is positioned vertically. It is actively emitting a vibrant, opaque red smoke trail from its rear.} The surface appears smooth and metallic.
\\
\end{tcolorbox}
}
\caption{An example of Discriminative\_State Subtask.}
\label{tab:dis_sta}
\end{table}

\begin{table}[t]
\small
\centering
\scalebox{1}{

\begin{tcolorbox}[colback=gray!00,%
                  colframe=black,%
                  width=8.5cm,%
                  arc=1.5mm, auto outer arc,
                  breakable,
                  left=0.9mm, right=0.9mm,
                  boxrule=0.9pt, colbacktitle = black!65!black,
                  title = {Subcategory 5: Spatial\_Relationship}
                 ]
{\textbf{Definition:} Grounding the target based on its spatial position relative to other entities.} \\
\begin{center}
\includegraphics[width = 0.8\linewidth]{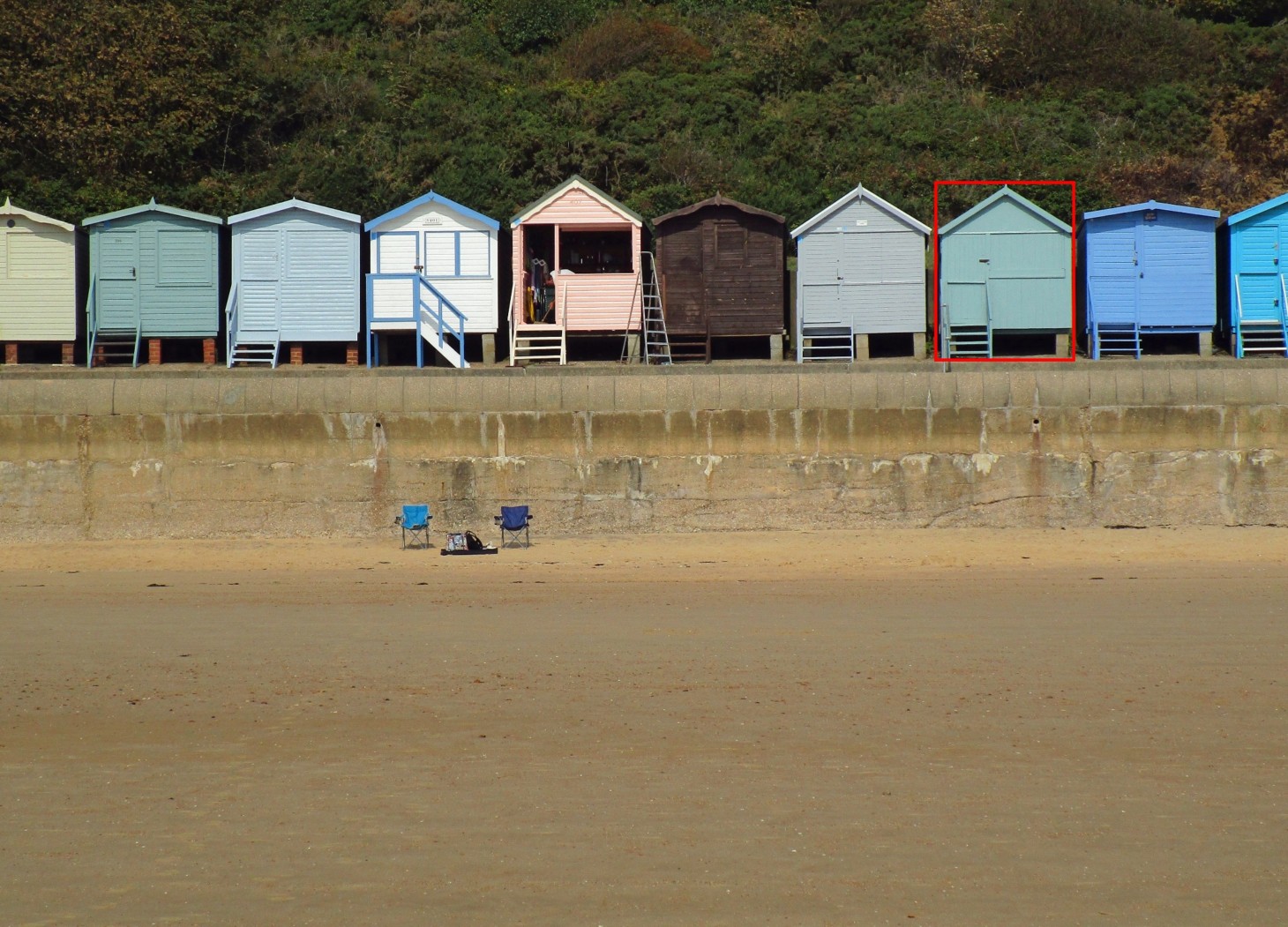}
\end{center}
\textbf{Description:} 
\\The object is a hut. \textbf{Immediately to the left of this hut is another beach hut, which is light grey in color. To its immediate right is a vibrant blue beach hut.} Below the hut is a sturdy concrete wall or barrier, and further down is the sandy beach. Directly above and behind the hut is a lush green hillside covered with dense vegetation.
\\
\end{tcolorbox}
}
\caption{An example of Spatial\_Relationship Subtask.}
\label{tab:spa_rel}
\end{table}

\begin{table}[t]
\small
\centering
\scalebox{1}{

\begin{tcolorbox}[colback=gray!00,%
                  colframe=black,%
                  width=8.5cm,%
                  arc=1.5mm, auto outer arc,
                  breakable,
                  left=0.9mm, right=0.9mm,
                  boxrule=0.9pt, colbacktitle = black!65!black,
                  title = {Subcategory 6: Spatial\_Counting}
                 ]
{\textbf{Definition:} Grounding the target based on explicit quantitative or ordinal information within the scene.} \\
\begin{center}
\includegraphics[width = 0.8\linewidth]{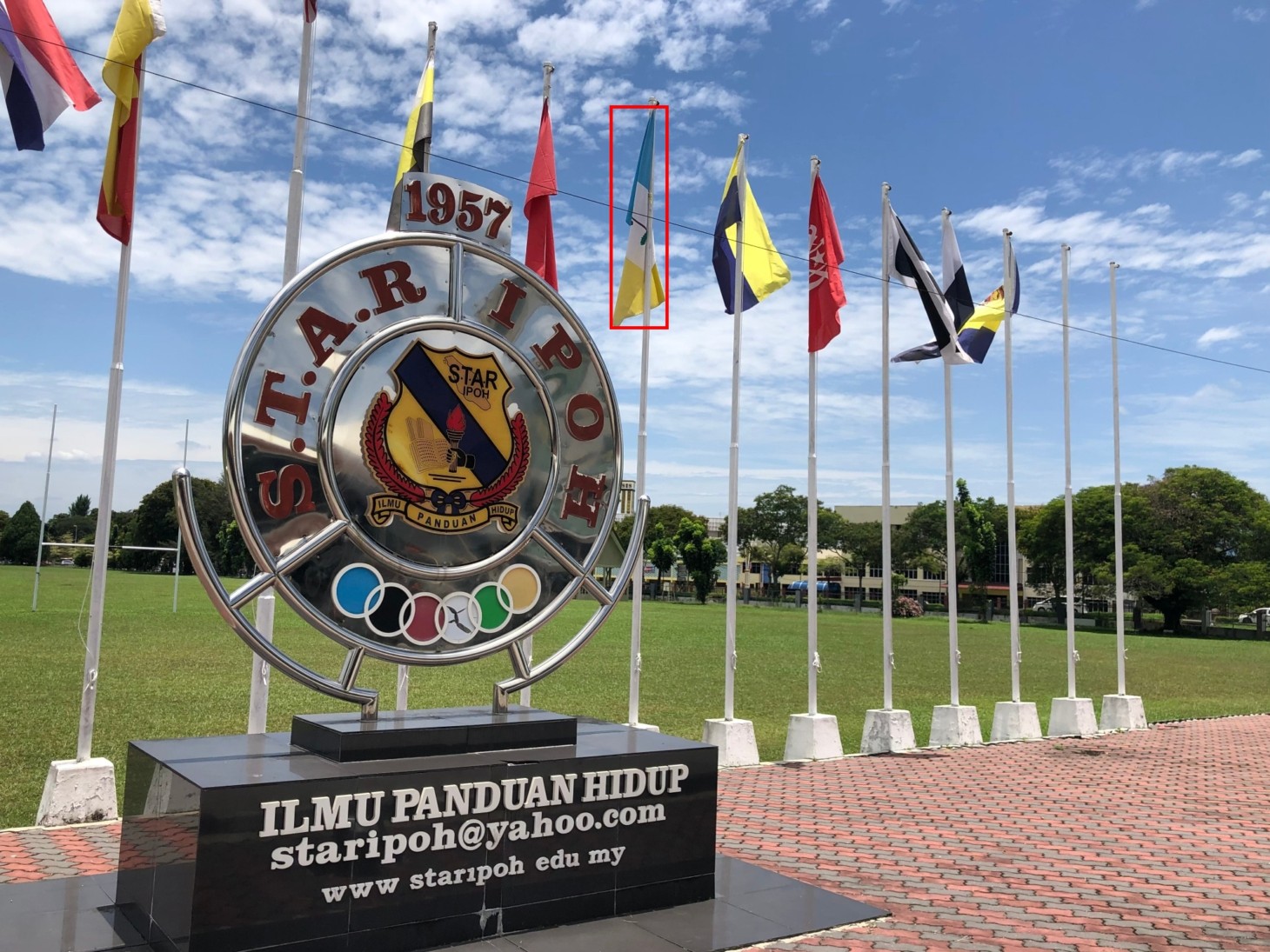}
\end{center}
\textbf{Description:} 
\\The flag is made of fabric with creases as it moves in the breeze. \textbf{To its right, there are five more flags.}
\\
\end{tcolorbox}
}
\caption{An example of Spatial\_Counting Subtask.}
\label{tab:spa_cnt}
\end{table}

\begin{table}[t]
\small
\centering
\scalebox{1}{

\begin{tcolorbox}[colback=gray!00,%
                  colframe=black,%
                  width=8.5cm,%
                  arc=1.5mm, auto outer arc,
                  breakable,
                  left=0.9mm, right=0.9mm,
                  boxrule=0.9pt, colbacktitle = black!65!black,
                  title = {Subcategory 7: Limited\_Occlusion}
                 ]
{\textbf{Definition:} Localizing objects with partial visibility caused by occlusion or truncation by the image frame.} \\
\begin{center}
\includegraphics[width = 0.8\linewidth]{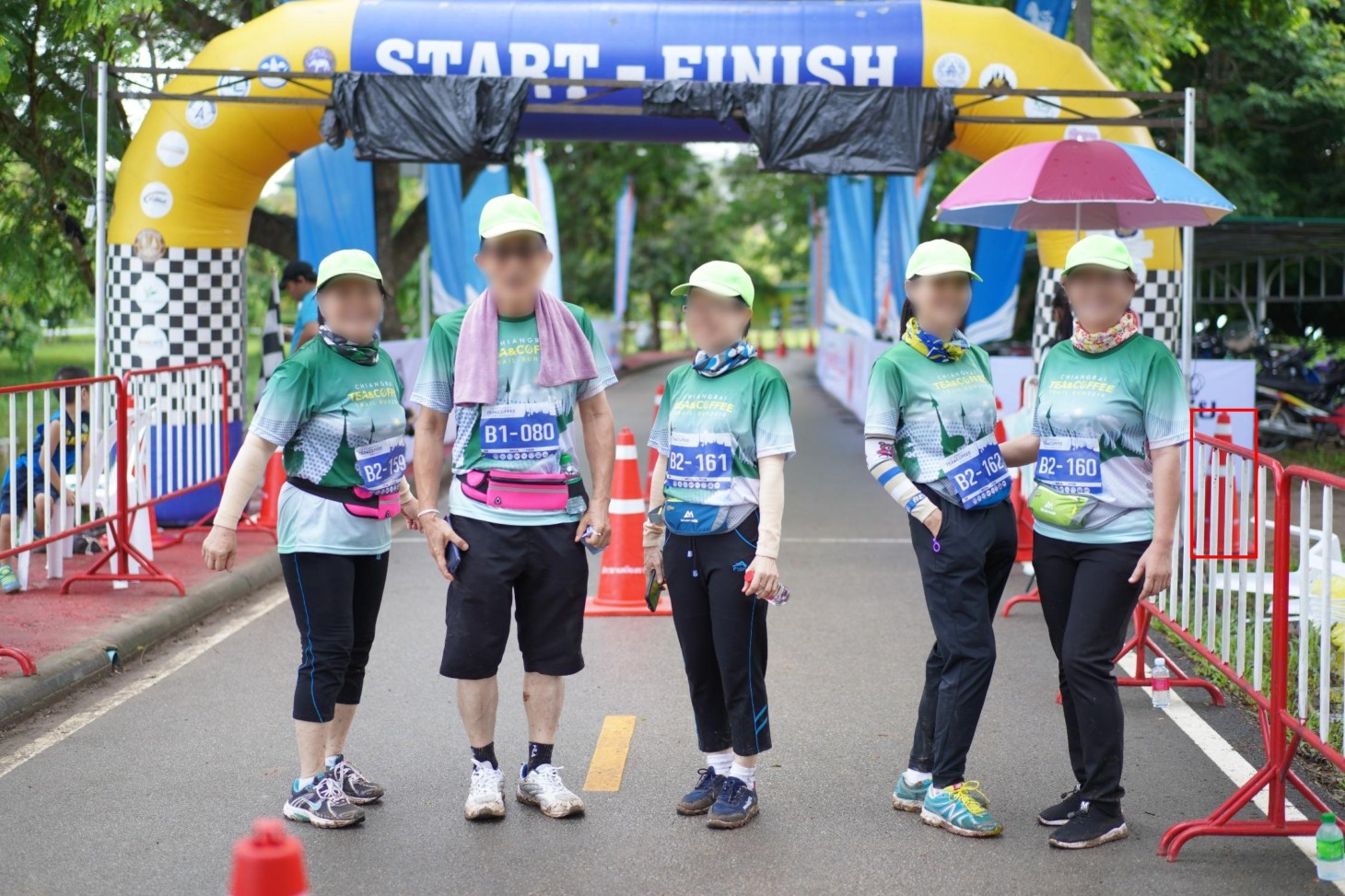}
\end{center}
\textbf{Description:} 
\\The object is a traffic cone, \textbf{the first one from the right.}
\\
\end{tcolorbox}
}
\caption{An example of Limited\_Occlusion Subtask.}
\label{tab:lim_occ}
\end{table}

\begin{table}[t]
\small
\centering
\scalebox{1}{

\begin{tcolorbox}[colback=gray!00,%
                  colframe=black,%
                  width=8.5cm,%
                  arc=1.5mm, auto outer arc,
                  breakable,
                  left=0.9mm, right=0.9mm,
                  boxrule=0.9pt, colbacktitle = black!65!black,
                  title = {Subcategory 8: Limited\_Small}
                 ]
{\textbf{Definition:} Localizing objects with diminutive scale in ultra-high resolution images.} \\
\begin{center}
\includegraphics[width = 0.8\linewidth]{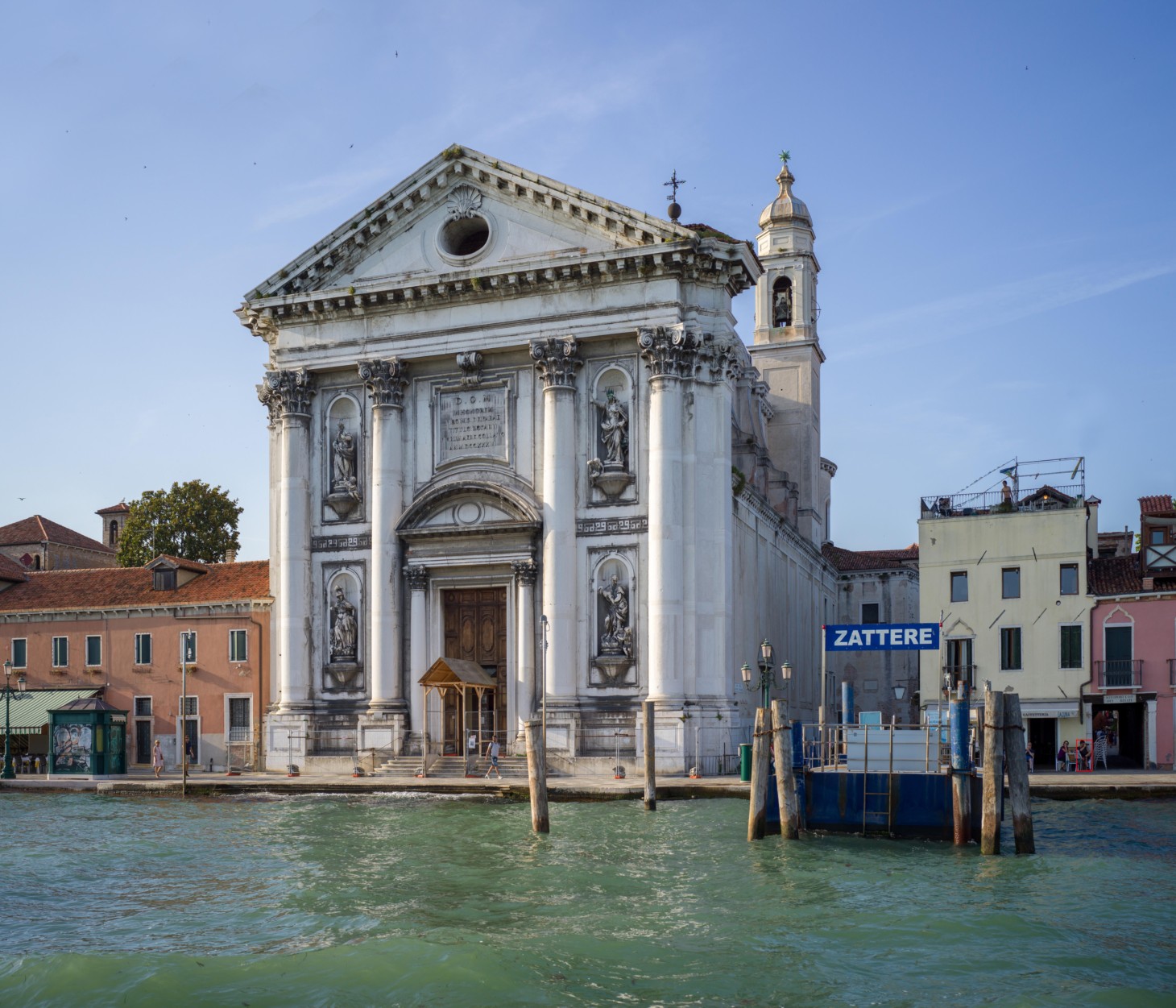}
\end{center}
\textbf{Description:} 
\\\textbf{The object is a person holding a camera.} It appears to be capturing a photograph while seated outside.
\\
\end{tcolorbox}
}
\caption{An example of Limited\_Small Subtask.}
\label{tab:lim_sml}
\end{table}

\begin{table}[t]
\small
\centering
\scalebox{1}{

\begin{tcolorbox}[colback=gray!00,%
                  colframe=black,%
                  width=8.5cm,%
                  arc=1.5mm, auto outer arc,
                  breakable,
                  left=0.9mm, right=0.9mm,
                  boxrule=0.9pt, colbacktitle = black!65!black,
                  title = {Subcategory 9: Rejection\_Appearance}
                 ]
{\textbf{Definition:} Rejecting the query due to a factual contradiction in the described visual attributes like color or texture.} \\
\begin{center}
\includegraphics[width = 0.8\linewidth]{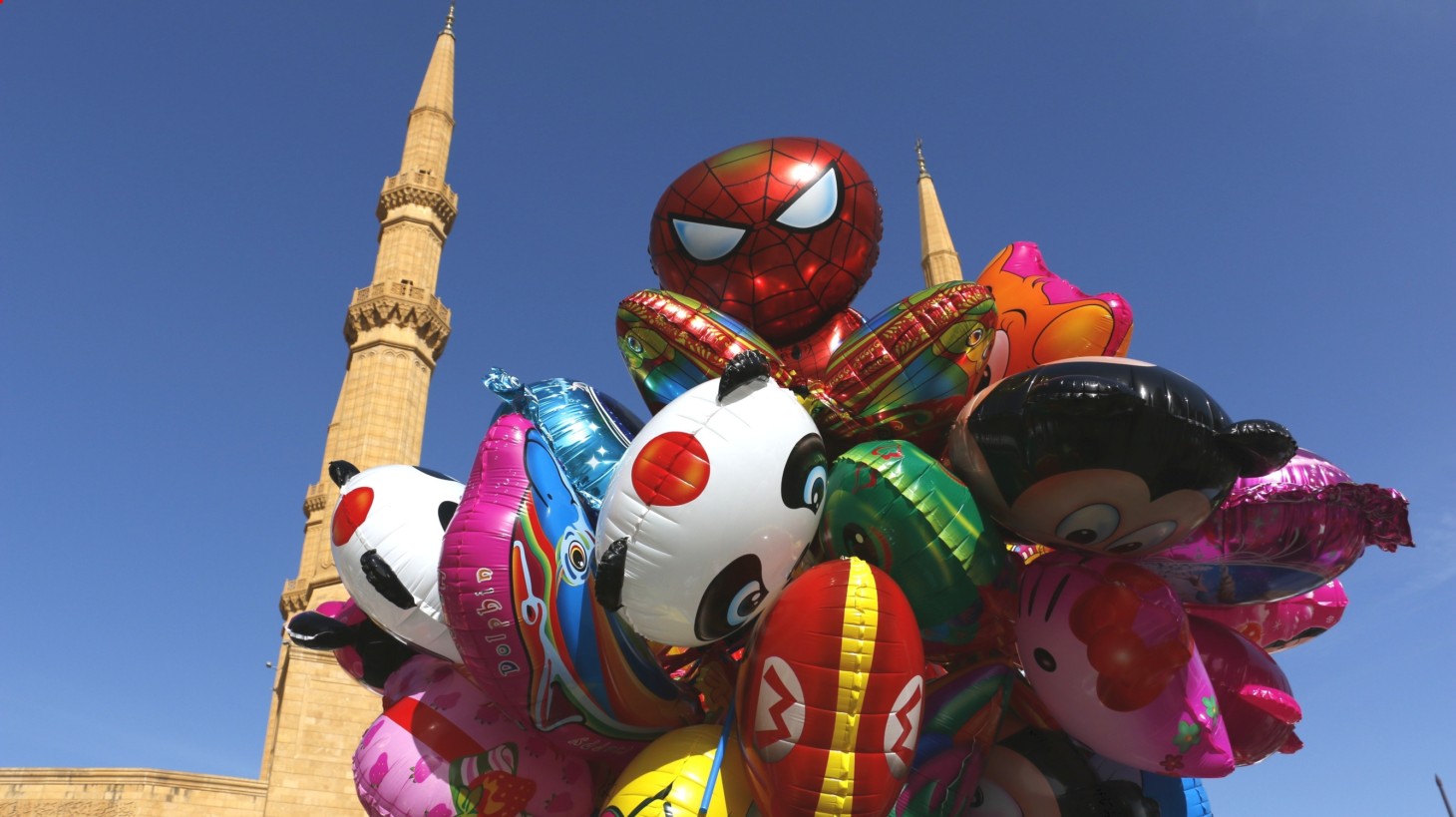}
\end{center}
\textbf{Description:} 
\\The object is an elongated, oval-shaped foil balloon, \textbf{primarily red with a prominent vertical white stripe running down its center. On either side of the white stripe, there are yellow, circle shapes outlined with red patterns.} The balloon has a smooth, reflective texture typical of Mylar balloons.
\\
\end{tcolorbox}
}
\caption{An example of Rejection\_Appearance Subtask.}
\label{tab:rej_app}
\end{table}

\begin{table}[t]
\small
\centering
\scalebox{1}{

\begin{tcolorbox}[colback=gray!00,%
                  colframe=black,%
                  width=8.5cm,%
                  arc=1.5mm, auto outer arc,
                  breakable,
                  left=0.9mm, right=0.9mm,
                  boxrule=0.9pt, colbacktitle = black!65!black,
                  title = {Subcategory 10: Rejection\_Component}
                 ]
{\textbf{Definition:} Rejecting the query due to a factual contradiction concerning a specific structural component.} \\
\begin{center}
\includegraphics[width = 0.8\linewidth]{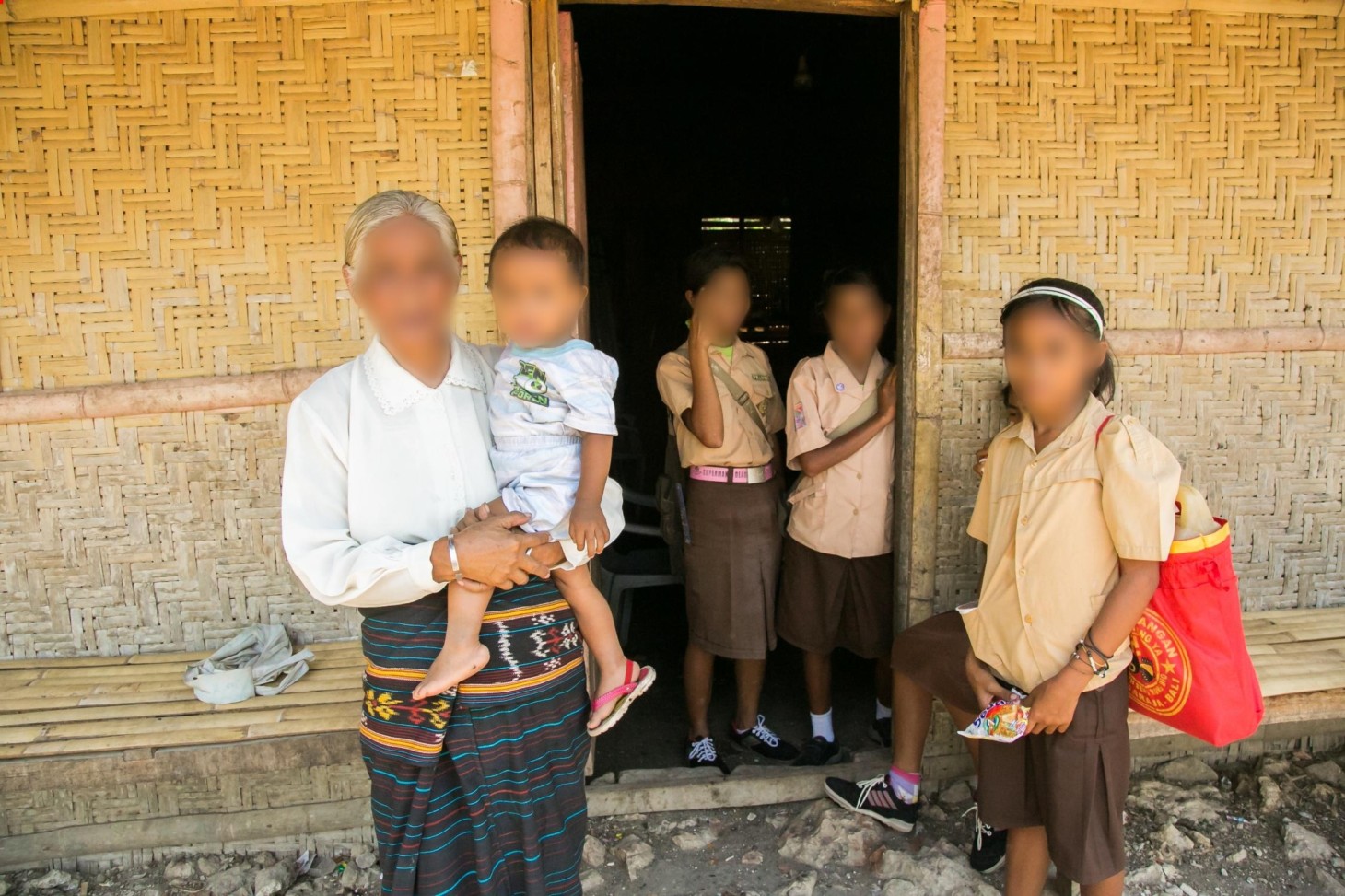}
\end{center}
\textbf{Description:} 
\\The object is a young child, consisting of their bare legs and small feet. \textbf{The child wears a pair of pink flip-flop with a white sole on both feet.}
\\
\end{tcolorbox}
}
\caption{An example of Rejection\_Component Subtask.}
\label{tab:rej_cmp}
\end{table}

\begin{table}[t]
\small
\centering
\scalebox{1}{

\begin{tcolorbox}[colback=gray!00,%
                  colframe=black,%
                  width=8.5cm,%
                  arc=1.5mm, auto outer arc,
                  breakable,
                  left=0.9mm, right=0.9mm,
                  boxrule=0.9pt, colbacktitle = black!65!black,
                  title = {Subcategory 11: Rejection\_Text}
                 ]
{\textbf{Definition:} Rejecting the query due to a factual mismatch with embedded textual information.} \\
\begin{center}
\includegraphics[width = 0.8\linewidth]{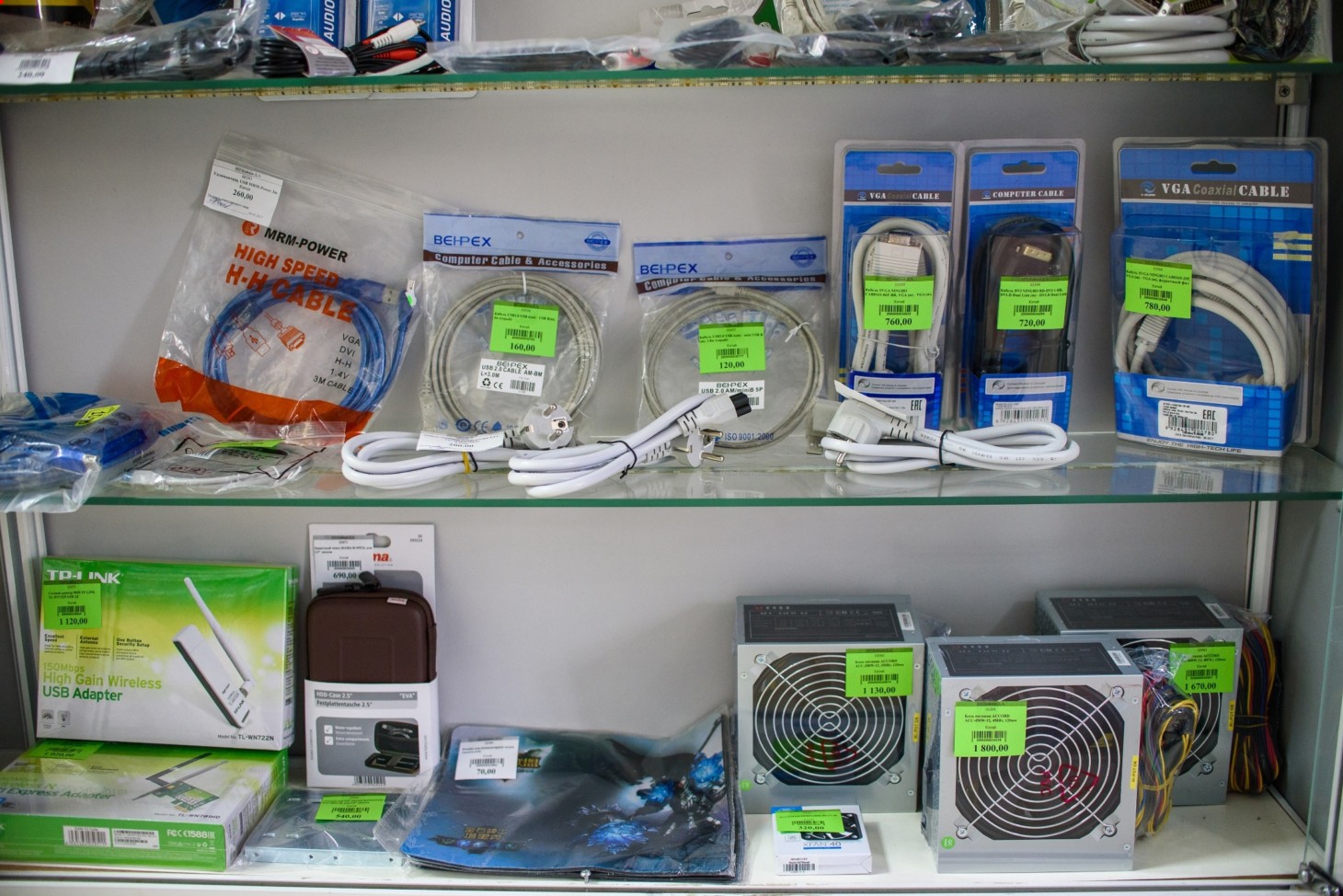}
\end{center}
\textbf{Description:} 
\\The object is a white VGA coaxial cable coiled inside a clear plastic blister pack with a blue backing. \textbf{A green price label with black text "180" is affixed to the front of the packaging. }The cable itself has a smooth appearance and is neatly coiled into a circular shape.
\\
\end{tcolorbox}
}
\caption{An example of Rejection\_Text Subtask.}
\label{tab:rej_txt}
\end{table}

\begin{table}[t]
\small
\centering
\scalebox{1}{

\begin{tcolorbox}[colback=gray!00,%
                  colframe=black,%
                  width=8.5cm,%
                  arc=1.5mm, auto outer arc,
                  breakable,
                  left=0.9mm, right=0.9mm,
                  boxrule=0.9pt, colbacktitle = black!65!black,
                  title = {Subcategory 12: Rejection\_State}
                 ]
{\textbf{Definition:} Rejecting the query because the object's described dynamic or static condition is factually incorrect.} \\
\begin{center}
\includegraphics[width = 0.8\linewidth]{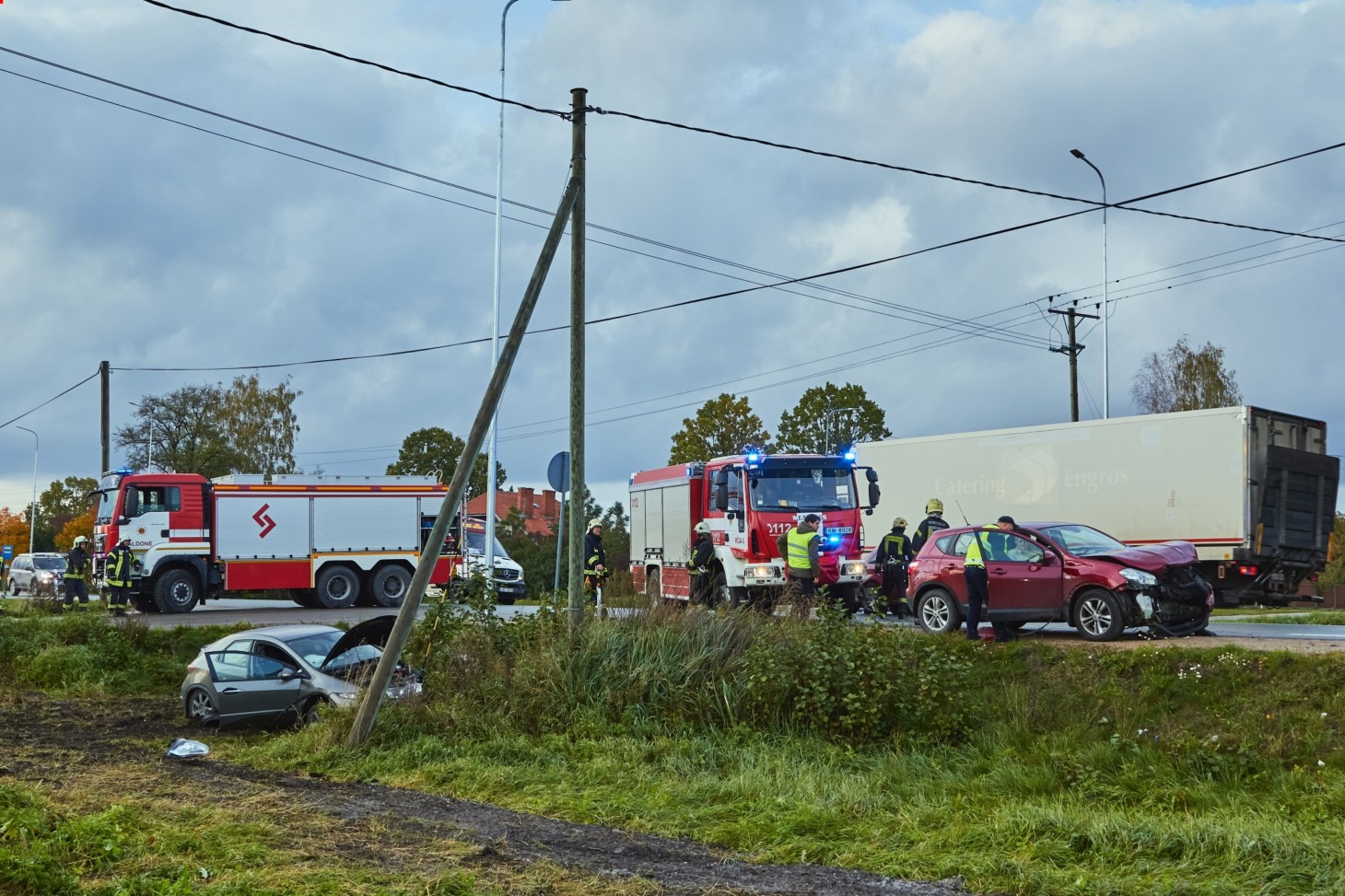}
\end{center}
\textbf{Description:} 
\\\textbf{The object is a maroon-colored compact SUV with its doors closed.} Its front end is heavily damaged and crushed, indicating an impact. The paint on the undamaged parts of the car appears somewhat glossy, and the windshield and windows are visible.
\\
\end{tcolorbox}
}
\caption{An example of Rejection\_State Subtask.}
\label{tab:rej_sta}
\end{table}